\documentclass[nojss]{jss} 

\author{Felix L. Rios\\University of Basel \And Giusi Moffa\\University of Basel \And
        Jack Kuipers\\ETH Zürich}
\title{Benchpress: A Scalable and Versatile Workflow for Benchmarking Structure Learning Algorithms}

\Plainauthor{ } 
\Plaintitle{\pkg{Benchpress}: A Scalable and Versatile Workflow for Benchmarking Structure Learning Algorithms for Graphical Models} 
\Shorttitle{\pkg{Benchpress}: a scalable and versatile workflow for structure learning} 

\Abstract{
  Describing the relationship between the variables in a study domain and modelling the data generating mechanism is a fundamental problem in many empirical sciences. Probabilistic graphical models are one common approach to tackle the problem. 
Learning the graphical structure for such models is computationally challenging and a fervent area of current research with a plethora of algorithms being developed. 
To facilitate the benchmarking of different methods, we present a novel \smk workflow, called \ttl for producing  scalable, reproducible, and platform-independent benchmarks of structure learning algorithms for probabilistic graphical models.
\ttl is interfaced via a simple \proglang{JSON}-file, which makes it accessible for all users, while the code is designed in a fully modular fashion to enable researchers to contribute additional methodologies.
\ttl currently provides an interface to a large number of \emph{state-of-the-art} algorithms from libraries such as \pkg{BDgraph}, \pkg{BiDAG}, \pkg{bnlearn}, \pkg{causal-learn}, \pkg{gCastle}, \pkg{GOBNILP}, \pkg{pcalg}, \pkg{r.blip},  \pkg{scikit-learn}, \pkg{TETRAD},  and \pkg{trilearn} as well as a variety of methods for data generating models and performance evaluation. 
Alongside user-defined models and randomly generated datasets, the workflow also includes a number of standard datasets and graphical models from the literature, which may be included in a benchmarking study.
We demonstrate the applicability of this workflow for learning Bayesian networks in five typical data scenarios. 
The source code and documentation is publicly available from \path{https://benchpressdocs.readthedocs.io}.
 
}
\Plainkeywords{reproducibility, scalable benchmarking, probabilistic graphical models} 

\Address{
  Felix L. Rios\\
  Department of Mathematics and Computer Science\\
  University of Basel\\
  Spiegelgasse 1, 4051 Basel, Switzerland\\
  E-mail: \email{felix.leopoldo.rios@gmail.com}
\\\\
  Giusi Moffa\\
  Department of Mathematics and Computer Science\\
  University of Basel\\
  Spiegelgasse 1, 4051 Basel, Switzerland\\
  E-mail: \email{giusi.moffa@unibas.ch}
\\\\
Jack Kuipers\\
Department of Biosystems Science and Engineering\\
ETH Zürich\\
Mattenstrasse 26, 4058 Basel, Switzerland\\
E-mail: \email{jack.kuipers@bsse.ethz.ch}
}

\usepackage[frozencache=true,cachedir=mycache]{minted} 

\usepackage{amsmath}
\usepackage{listings}
\usepackage{subcaption}
\usepackage{xargs}
\usepackage{enumitem}
\usepackage{fancyvrb}

\usepackage{breakcites}
\usepackage[vertfit]{breakurl} 
\usepackage{microtype}

\usepackage[textwidth=4cm, textsize=footnotesize]{todonotes}
\usepackage{subfiles} 
\usepackage{xspace}
\usepackage[overload]{textcase}

\usepackage{tikz}
\usepackage{dirtytalk}
\usetikzlibrary{shapes,arrows}
\usepackage{hyperref}

\newcommand{\ttl}{\pkg{Benchpress}\xspace}
\newcommand{\smk}{\pkg{Snakemake}\xspace}
\newcommand{\sng}{\pkg{Apptainer}\xspace}
\newcommand{\dkr}{\pkg{Docker}\xspace}

\newcommand\hmm[1]{\ifnum\ifhmode\spacefactor\else2000\fi>1000 \uppercase{#1}\else#1\fi}

\lstdefinelanguage{json}{
    basicstyle=\normalfont\ttfamily,
    numbers=left,
    numberstyle=\scriptsize,
    stepnumber=1,
    numbersep=8pt,
    showstringspaces=false,
    breaklines=true,
    frame=lines,
    backgroundcolor=\color{background},
    literate=
     *{0}{{{\color{numb}0}}}{1}
      {1}{{{\color{numb}1}}}{1}
      {2}{{{\color{numb}2}}}{1}
      {3}{{{\color{numb}3}}}{1}
      {4}{{{\color{numb}4}}}{1}
      {5}{{{\color{numb}5}}}{1}
      {6}{{{\color{numb}6}}}{1}
      {7}{{{\color{numb}7}}}{1}
      {8}{{{\color{numb}8}}}{1}
      {9}{{{\color{numb}9}}}{1}
      {:}{{{\color{punct}{:}}}}{1}
      {,}{{{\color{punct}{,}}}}{1}
      {\{}{{{\color{delim}{\{}}}}{1}
      {\}}{{{\color{delim}{\}}}}}{1}
      {[}{{{\color{delim}{[}}}}{1}
      {]}{{{\color{delim}{]}}}}{1},
}

\newcommand\ci{\perp\!\!\!\perp}
\newcommand{\clique}{C}
\newcommand{\cliquesp}{\mathcal{C}}
\newcommand{\covmat}{\Sigma}
\newcommandx{\dagsp}[1][1=]{\ifthenelse{\equal{#1}{}}{\mathcal D}{\mathcal D_{#1}}}
\newcommand{\dens}{f}

\newcommandx{\dgraphsp}[1][1=]{\ifthenelse{\equal{#1}{}}{\mathcal U}{\mathcal U_{#1}}}
\newcommand{\distr}{\Prob}

\newcommand{\felix}[1]{}

\newcommand{\genseta}{A}
\newcommand{\gensetb}{B}
\newcommand{\gensetc}{C}

\newcommandx{\graph}[1][1=]{
    \ifthenelse{\equal{#1}{}}
        {G}
        {G_{#1}}
}
\newcommandx{\graphb}[1][1=]{
    \ifthenelse{\equal{#1}{}}
        {G'}
        {G_{#1}'}
}
\newcommandx{\graphsp}[1][1=]{\ifthenelse{\equal{#1}{}}{\mathcal G}{\mathcal G_{#1}}}

\newcommand{\graphnodeset}{V}
\newcommand{\graphedgeset}{E}

\newcommand{\iid}{\texttt{iid}}

\newcommand{\kernel}[1]{\mathbf{#1}}
\newcommand{\nalgs}{54\xspace}

\newcommand{\p}{p}
\newcommand{\pa}{pa}
\newcommandx{\pasp}[1][1=]{
\ifthenelse{
\equal{#1}{}}
    {\mathcal{Pa}}
    {\mathcal Pa(#1)}
}
\newcommand{\param}{\Theta}

\newcommand{\probf}{pr}
\newcommand{\randbn}{\texttt{bin\_bn}}
\newcommand{\randdag}{\texttt{pcalg\_randdag}}
\newcommand{\rma}{\mathbf Y}
\newcommand{\rva}{Y}

\newcommand{\rvc}{Z}
\newcommand{\sep}{S}
\newcommand{\seps}{\mathcal{S}}

\newcommandx{\trsp}[1][1=]{\ifthenelse{\equal{#1}{}}{\mathcal T}{\mathcal T_{#1}}}

\newcommandx{\uk}[2][1=]{
\ifthenelse{
\equal{#1}{}}
    {\kernel{Q}_{#2}}
    {\kernel{Q}_{#2} \langle #1 \rangle}
}
\newcommand{\va}{y}
\newcommand{\vals}[1]{\code{#1}} 

\begin{document}

\section{Introduction}
Probabilistic graphical models play a central role in modern statistical data analysis.
Their compact and elegant way of visualising and representing complex dependence structures in multivariate probability distributions has shown to be successfully applicable in many scientific domains, ranging from disciplines such as social sciences and image analysis to biology, medical diagnosis and epidemiology \citep[see \emph{e.g.},][]{elwert2013graphical, friedman2000using, friedman2004inferring,10.1093/schbul/sbx013, kuipers2018mutational,  kuipers_moffa_kuipers_freeman_bebbington_2019}.
\felix{Intro to and application of GMs.}

One of the main advantages of graphical models is that they provide a tool for experts and researchers from non-statistical fields to easily specify their assumptions in a given problem domain, in a highly transparent way.
However, in many realistic situations, the number of variables is either too large to build networks by hand, as it is for example the case in genomic applications, or simply no prior knowledge is available about the relationships between different variables.
As a consequence, there has been a growing interest in automated strategies to infer the graph component of a probabilistic graphical model from data, so-called \emph{structure learning} or \emph{causal discovery}. 
For recent reviews of the wide flora of algorithms that have emerged in the last two decades, the reader is referred to \cite{koski2012review,Kitson2023survey}.

\felix{Intro to structure learning.}
Since the scientific principle of reproducibility is instrumental for high-quality standards in research, there is an increasing demand to present new algorithms with publicly available source code and datasets \citep[see \emph{e.g.},][]{Munafo:2017tp, Lamprecht:2020uj}. 
With publicly available software we should, in principle, be able to easily compare different methods with respect to their performance in different settings. The reality is that the practical implementation of new approaches displays high levels of heterogeneity in many aspects, which renders comparative studies both challenging and time-consuming.
One of the main difficulties is that different algorithms may be implemented in different programming languages, with different dependencies. 
Different packages may differ with respect to the representation of graphical models, the data formats, and the way they interface to the user, \emph{e.g.},\ through the command-line arguments, a configuration file, or a function in a specific programming language.
Another complication is that the number of computations increases rapidly in comparison studies, especially when input parameters are altered, which requires parallel computation capabilities and careful bookkeeping to record and organise the results for reporting.
As a result, researchers may invest considerable effort in producing benchmarking scripts, and may only be able to implement comprehensive comparisons for a handful of relevant algorithms.
To complicate matters further, there is no single well-established metric to evaluate the performance of structure learning algorithms. 
On the contrary, there is a wide range of different metrics to choose from \citep[see \emph{e.g.},][]{tsamardinos2006max, scutari2019learns, constantinou2021large} and researchers tend to use only a small selected subset of these to evaluate a new algorithm, typically focusing on those relevant to the problem their algorithm addresses. 

When it comes to datasets, there are standard choices commonly used in the literature for benchmarking new algorithms. While desirable to also reduce the researchers' degrees of freedom in terms of data choices, focusing on a limited set of data may have damaging effects by unintentionally leading to solutions specifically tailored to the data properties, rather than more generally targeting structure learning. 
Despite the ongoing rapid advances in causal discovery algorithms, there is still a lack of unified approaches for simulating benchmarking data that display realistic features and are easily accessible to the user. A recent effort to address this need aims to use assembly lines' data for generating semisynthetic manufacturing data with known causal relationships \citep{gobler2023textttcausalassembly}.
\cite{montagna2023assumption} focus on data generated from models violating the standard assumptions for causal discovery.


\felix{Identify the problem of comparing }
With the problems described above in mind, the objective of this work is to develop a unified framework to facilitate the execution and benchmarking of different algorithms. 
We present a novel \smk \citep{10.1093/bioinformatics/bts480} workflow called \ttl, which is designed for reproducible, platform-independent, and fully scalable benchmarks of structure learning algorithms.
\ttl is interfaced via an easy-to-use configuration file in \proglang{JSON}-format \citep{pezoa2016foundations},
where a benchmark setup is specified in a module-based \emph{probabilistic programming} style, separating model specification, in terms of graph and corresponding parameters, from algorithm execution and evaluation.
\smk enables \ttl to seamlessly scale the computations to server, cluster, grid and cloud environments, without any changes in the configuration file.
The support for containerized software through \emph{e.g.}, \sng \citep{kurtzer2017singularity} or \dkr \citep{merkel2014docker} images, together with platform-independent representations of data sets and graphs enables \ttl to compare algorithms from different libraries, possibly implemented in different programming languages and with different dependencies.
Creating a consensus benchmark setup may mitigate distortions in the performance comparison inadvertently resulting from researchers' degrees of freedom. At the same time, the possibility of easily designing reproducible benchmarking settings will encourage researchers to conduct more comprehensive comparison studies on different datasets and using different metrics.

\ttl is implemented in a modular coding style with functionality for researchers to contribute with additional modules for generating models, structure learning algorithms and evaluating results. 
In the current publicly released version we have already included \nalgs algorithm modules from some of the most popular libraries such as \pkg{BDgraph} \citep{JSSv089i03}, \pkg{BiDAG} \citep{suter2021bayesian}, \pkg{bnlearn} \citep{JSSv035i03}, \pkg{causal-learn} \citep{causallearn}, \pkg{gCastle} \citep{zhang2021gcastle},  \pkg{GOBNILP} \citep{cussens2020gobnilp}, \pkg{pcalg} \citep{kalisch2012causal}, \pkg{r.blip} \citep{scanagatta2015learning}, \pkg{scikit-learn} \citep{scikit-learn}, \pkg{TETRAD} \citep{glymour1986causal}, and \pkg{trilearn} \citep{olsson2019} along with models and data sets from the standard literature.
See Table~\ref{tab:algorithms} in the Appendix for a complete list of the currently available algorithms.

\felix{The solution}

With comparable intent, \cite{constantinou2020bayesys} developed \pkg{Bayesys},  albeit exclusively designed for \emph{Bayesian networks}, a specific type of probabilistic graphical models \citep{pearl1997probabilistic}. 
Furthermore \pkg{Bayesys} is a \proglang{Java} implementation only including six algorithms for structure learning at the time of writing. 
\felix{Relate to other attempts to solve the same problem.}
Another tool called \pkg{causal-compare} with a similar purpose to \ttl was released as part of the \pkg{TETRAD} project \citep{JMLR:v21:19-773}.
However, the functionality of \pkg{causal-compare} is restricted to the algorithm implementations in the \pkg{TETRAD} project, representing only a subset of available algorithms from the literature.
Overall \ttl offers wider scope and usability than currently available alternatives since \ttl only requires \dkr, without placing any special requirements on the implementation of individual algorithms. 

\ttl is distributed under the GPL v2.0 License, encouraging its use in both, academic and commercial settings. 
Source code and documentation are available from \url{https://benchpressdocs.readthedocs.io}.

The rest of the paper is structured as follows.
Section~\ref{sec:pgms} presents an introduction to graphical models along with some notational conventions.
Section~\ref{sec:structure_learning} provides background on current strategies for structure learning.
Section~\ref{sec:json} describes the structure of the \proglang{JSON} interface and the modules already available for benchmarking.
In Section~\ref{sec:case_study}, we show how to use \ttl in several typical benchmarking scenarios.
Installation and usage guidelines are provided in Section~\ref{sec:installation}, while Section~\ref{sec:newalg} discusses how to add new algorithms to the \ttl framework.
\felix{Paper structure}

\section{Probabilistic graphical models}\label{sec:pgms}
This section provides a brief introduction to graphical models and graph theory, with definitions as in  \emph{e.g.}, \cite{diestel2005graph, lauritzen1996}.
Let \(\distr\) be a \(\p\)-dimensional distribution and let \(\rva := (\rva_i)_{i=1}^p\) be a random vector such that \(\rva \sim \distr\).
Further let  $\graph=(\graphnodeset, \graphedgeset)$ be a graph, where  \(\graphnodeset=\{\rva_i\}_{i=1}^\p\) is the node set and \(\graphedgeset\) is the edge set, consisting of ordered pairs of distinct elements of \(\graphnodeset\).
Then \(\distr\), or alternatively \(\rva\), is said to be Markov with respect to \(\graph\) if separation, based on some graphical criteria, of two subsets of nodes,  $\rva_\genseta$ and $\rva_\gensetb$, by a third set $\rva_\gensetc$ in \(\graph\) implies their conditional independence in \(\distr\), where $\rva_\genseta := (\rva_i)_{i\in \genseta}$.
The term probabilistic graphical model or graphical model has been used interchangeably in the literature to refer to either the tuple \((\graph, \distr)\) or a collection of distributions \( \mathcal P_{\graph}\), usually restricted to some specific parametric family, where every \(\distr' \in \mathcal P_{\graph} \) is Markov with respect to \(\graph{}\).
In this text, we will use the former meaning.
In the case where \(\distr\) is restricted to a specific parametric family, it is usually uniquely determined by some parameter vector \(\param\), thus either notation may be used.
\felix{Define graphical models}

Assuming \(\dens \) is a density for \(\distr\), conditional independence between the random variables  $\rva_\genseta$ and $\rva_\gensetb$ given $\rva_\gensetc$ is well-defined as \(\dens(\va_\genseta|\va_\gensetb, \va_\gensetc) = \dens(\va_\genseta|\va_\gensetc),\) where \( (\va_\genseta,\va_\gensetb, \va_\gensetc) \in \mathcal\rva_\genseta 
\times \mathcal\rva_\gensetb \times \mathcal\rva_\gensetc \) and \(\mathcal\rva_\genseta\) denotes the range of \(\rva_\genseta\).
In contrast, node separation may have different definitions depending on the type of graph considered.
Typical classes of graphs are: \emph{undirected graphs}, where edges are specified as unordered tuples and separation of  $\rva_\genseta$ and $\rva_\gensetb$  given $\rva_\gensetc$ means that every path between $\rva_\genseta$ and $\rva_\gensetb$ must pass $\rva_\gensetc$; \emph{directed acyclic graphs} (DAGs), where edges are represented as ordered tuples and separation is defined by a concept known as \emph{d-separation}, see \emph{e.g.},  \cite{pearl1997probabilistic}.
Next follows a brief introduction to the type of graphical models currently implemented in \ttl.

\felix{Examples of graphs: Bayesian networks}
\emph{Bayesian networks} constitute one of the most popular classes of graphical models, referring to distributions that are Markov with respect to a DAG. 
The conditional independence encoded by the DAG implies that the density can be factorised into local conditional densities, where each node \(\rva_i\) is directly dependent on its \emph{parents}, \( \pa(\rva_i)\), as
\begin{align*}
    \dens(\va) = \prod_{i=1}^p \dens(\va_i | \pa(\va_i)),
\end{align*}
where \(\dens(\va_i | \pa(\va_i))\) denotes the conditional density of \(\rva_i\) given \(\pa(\rva_i)=\pa(\va_i)\).
This property is appealing for several reasons.
Firstly, the decomposition into local conditional probability distribution provides an intuitive way to express and visualise knowledge about a specific problem domain.
In particular, an expert could express a causal mechanism in terms of a DAG, which in turn provides a mathematical language to answer causal queries from non-experimental data \citep{pearl1995causal}.
Further, it enables fast computations of conditional and marginal distribution using message-passing algorithms by exploiting the factorisation property in the distribution.
However, when the DAG is not used for expressing causal knowledge, the direction of the edges may be misleading since the conditional independencies of a distribution are not uniquely encoded by a single DAG in general but rather by a class of \emph{Markov equivalent} DAGs, all of which encode the same conditional independence statements.
One way to represent a Markov equivalence class is by a \emph{completed partially directed acyclic graph} \citep[CPDAG, see][]{chickering2013transformational}. 
A CPDAG is a partially directed DAG which can be obtained from a DAG by regarding an edge \((\rva_i,\rva_j)\) as undirected if and only if both directed edges \((\rva_i,\rva_j)\) and \((\rva_j,\rva_i)\) are present in some of the DAGs in the same equivalence class.
The CPDAG is also sometimes called the \emph{essential graph}.
An equivalent way to represent the conditional independence statements in a DAG is through the \emph{pattern graph}. 
The pattern graph of a DAG is the graph obtained by keeping the directions of the \emph{v-structures} and regarding the remaining edges as undirected. 
v-structures are characterised by triples of nodes where the edges of two of the nodes, which should not be neighbours, are pointing to the third. 

\felix{Introduce Markov networks}

\emph{Markov networks} or \emph{Markov random fields} are another popular class of graphical models, referring to distributions that are Markov with respect to an undirected graph.
There is, in general, no decomposition into local densities for the models in this class.
However, there exist functions \(\{\phi_{\clique}\}_{\clique\in \cliquesp}\) such that
\begin{align*}
        \dens(\va) = \prod_{\clique \in \cliquesp} \phi_{\clique}(\va_\clique),
\end{align*}
where \(\cliquesp\) is the set of maximal subgraphs where all nodes are directly connected with each other, usually called \emph{cliques}, which may be leveraged to calculate marginal distributions efficiently.
There is a sub-class of undirected graphs called \emph{decomposable} (DG) or \emph{chordal} or \emph{triangulated}, where each cycle with more than three nodes has a \emph{chord}, an edge which breaks it.
Distributions which are Markov with respect to decomposable graphs are called \emph{decomposable} and their densities are built by local, clique-specific densities, \(\{\dens_\clique\}_{\clique \in \cliquesp}\) and their marginals so that the joint density factorises as
\begin{align*}
    \dens(\va) = \frac{\prod_{\clique \in \cliquesp} \dens_\clique(\va_{\clique})}{\prod_{\sep \in \seps} \dens_\sep(\va_\sep)},
\end{align*}
where \(\seps\) is the multi-set of separators in \(\graph\), found \emph{e.g.}, as the intersection of neighboring cliques in a so-called \emph{junction tree}, built from the cliques in the graph; see \emph{e.g.}, \cite{lauritzen1996} for details.
In general Markov networks and Bayesian networks encode different sets of conditional independence statements, and there are sets which we can represent with one model but not the other. 
Decomposable graphs represent the intersection of those sets of statements which we can represent by both undirected graphs and DAGs \citep[see \emph{e.g.},][]{lauritzen1996, cowell2003probabilistic, koller2009probabilistic}.
The DAGs representing the intersection set are called \emph{perfect}, meaning that all the parents of each node are connected by an edge.
Further characterisations of decomposable graphs are found in \emph{e.g.}, \cite{lauritzen1996, cowell2003probabilistic} and \cite{duarte2021new}.



\felix{Introduce Markov networks}

\felix{Summarize the above discussion}

\section{Structure learning}
Structure learning algorithms for Bayesian networks were proved to be NP-hard, even when the number of parents of each node is restricted to two \citep{chickering2004large}.
Also, the number of undirected graphs grows as \( \mathcal O(2^{p^2})\) even when restricted to decomposable graphs, making an exhaustive search for a specific graph infeasible \citep[see \emph{e.g.},][]{countchordal, cta,hebertjohnson2023counting}. 
Therefore, in practice, we must rely on approximation methods which we can divide into three main categories: \emph{constraint-based}, \emph{score-based}, and \emph{hybrid} algorithms, discussed briefly below.

\felix{Intro to structure learning}\label{sec:structure_learning}

The seed for the first constraint-based algorithm was planted by \cite{verma1991equivalence} in the context of causal Bayesian networks.
In line with their work, constraint-based methods employ conditional independence tests among the variables to first estimate an undirected graph by, \emph{e.g.}, starting with the complete graph, and excluding an edge \((\rva_i,\rva_j)\) if there exists some set \(\rva_\genseta\) such that the hypothesis \(\rva_i  \ci_{\distr} \rva_j \mid \rva_\genseta\) cannot be rejected according to a suitable test procedure. 
Such methods tend to be very fast but testing can suffer from large numbers of false negatives, \emph{i.e.}, falsely not including edges when they should be present. 
Relaxing the significance level of the tests to reduce false negatives can, however, rapidly inflate false positives and runtimes. 
\felix{Intro to constraint-based methods}

Score-based methods on the other hand aim to optimise a global score function defined on the graph space.
As a workaround to cope with the immense number of graphs, the search is often limited to certain types of graphs. 
For example in  DAG models, one may restrict the number of parents or combine DAGs into larger and more easily represented classes \citep[\emph{e.g.}, orders and partitions][]{Friedman2003, doi:10.1080/01621459.2015.1133426}. 
For decomposable graphs, a possibility is to limit the maximal clique size.
Score functions may rely on penalised log-likelihoods or on Bayesian posterior probabilities of graphical structures conditional on the observed data \citep[see \emph{e.g.},][]{koller2009probabilistic}. 
Besides mere optimisation, Bayesian structure learning also focuses on sampling procedure to provide a finer characterisation of the posterior graph distribution. 
When an expression for the posterior distribution is available at least up to a normalising constant, we can implement  Markov chain Monte Carlo (MCMC) schemes to sample from it, see \emph{e.g.}, \cite{madigan1995graphicalmodels, giudici2003improving, Friedman2003} for early works and \cite{ doi:10.1080/01621459.2015.1133426, jennings2018birth} for more recent strategies.
In Gaussian Markov networks, Monte Carlo sampling turns out to be infeasible for larger networks since there is in general no analytic expression for the un-normalised posterior \citep{Atay-Kayis01062005}.
On the other hand, score-based methods such as \emph{graphical lasso}, rely on the fact that the edges in the graph are revealed from the non-zero pattern of the elements in the precision (inverse covariance) matrix.
Optimisation techniques can successfully estimate the \(L_1\)-penalised log-likelihood of the precision matrix, and the graph as a by-product; see \cite{citeulike:2134265, Meinshausen2008880, doi:10.1198/jcgs.2011.11051a}.
Decomposable models, do possess a closed form expression for the un-normalised graph posterior and most inferential methods built on the representation via undirected graphs focus on Monte Carlo sampling \cite[see \emph{e.g.},][]{Giudici01121999, Green01032013, elmasri2017decomposable, olsson2019} with a few focusing on optimisation \cite[see \emph{e.g.},][]{carvalho2006structure, studeny2017towards, rantanen2017learning}.

\felix{Add something.. One example of decomposable scores is  hyper directed Markov property .}

\felix{Intro to score methods}

Hybrid algorithms typically use constraint-based algorithms as a preliminary step called \emph{restrict phase} to reduce the search space for a score-based method.
A score based-algorithm is then used in the \emph{maximise phase}  \citep[see \emph{e.g.},][]{tsamardinos2003algorithms, scanagatta2015learning,  doi:10.1080/10618600.2021.2020127}.

\felix{Intro to hybrids}

\felix{Summarize methods}

\section{Benchpress modules and JSON configuration files}\label{sec:json}

In this section, we describe the modules of \ttl and the structure of the \proglang{JSON} \citep{pezoa2016foundations} configuration file, which serves as the interface for the user.
As \ttl is implemented as a \smk workflow, a brief overview of the latter will help to illustrate the overall concept of the tool.
\smk is a highly popular workflow management system with a custom \proglang{Python}-based programming language enabling the so-called \smk \emph{rules}, which build the core of a workflow.
Each rule typically comprises four types of fields: the \code{input} and \code{output} fields, defining the input and output files, respectively, the \code{script} and \code{container} fields, defining the script for generating the output files based on the input files, and the \dkr image used for running the script, respectively. 

When requesting an output file from a specific rule, \smk will automatically pattern-match the input files of the rule with the output files of other rules, before executing the script. In this way, \smk will recursively start a chain reaction of executions, where each rule only depends on its ancestors through its input files. 
Thus, the structure of rule relationships may be represented as a DAG, where an arrow from rule A to rule B indicates that an output file from A matches an input file from B.
Each rule lives in an independent \sng container and the DAG relationship enables parallel executions.
Note that, \sng is a container management system which requires fewer operating system privileges than \emph{e.g.}\ \dkr, making it possible to run \ttl on shared servers as opposed to \dkr which is usually blocked for security reasons.


\smk has built-in support for reading configuration files in \proglang{JSON} format, passed as an argument to the program.
The \proglang{JSON} file provides a simple \emph{key-value} format which is easy to interpret and modify. 
This property is heavily exploited by \ttl, where the \proglang{JSON} file gives a complete interface for the user, without modifying the rules. 
When the value of a key is another \proglang{JSON} object or a list, we call it a \emph{section}.

\begin{figure}[tbp]
\centering
\tikzstyle{decision} = [diamond, draw, fill=blue!20, 
    text width=4.5em, text badly centered, node distance=3cm, inner sep=0pt]
\tikzstyle{block} = [rectangle, draw, fill=green!10, 
    text width=4em, text centered, rounded corners, minimum height=2em]
\tikzstyle{module} = [rectangle, draw, fill=blue!20, 
    text width=2.5em,  rounded corners, minimum height=2em]

\tikzstyle{line} = [draw, -latex']
\tikzstyle{altline} = [draw,fill=black, -latex']
\tikzstyle{uline} = [draw,fill=black, -latex']
\tikzstyle{cloud} = [draw, rectangle,fill=red!20, node distance=2.2cm, minimum height=1.5em]

\definecolor{data1}{RGB}{212, 126, 190}
\definecolor{data2}{RGB}{236, 185, 209}

\definecolor{stl1}{RGB}{60, 118, 175}
\definecolor{stl2}{RGB}{179, 199, 229}

\definecolor{eval1}{RGB}{138, 104, 180}
\definecolor{eval2}{RGB}{189, 173, 206}

\begin{tikzpicture}[node distance = 2cm, auto]
    
    \node [cloud, fill=data2] (graph) {$G_{k:m}$};
    \node [cloud, fill=data2, right of=graph] (param) {$\Theta_{k:m}$};
    \node [cloud, fill=data2, right of=param] (data) {$\mathbf Y_{k:m}$};

    \node [module, node distance=2cm, above of=graph, fill=data1] (gmod) {\texttt{graph}};
    \node [module, node distance=2cm, above of=param, text width=5.5em, fill=data1] (pmod) {\texttt{parameters}};
    \node [module, node distance=2cm, above of=data, fill=data1] (dmod) {\texttt{data}};
    
    \node [module, node distance=2cm, above of=dmod, text width=7.5em, fill=data1] (pmods) { \texttt{bdgraph\_rgwish}  \texttt{sem\_params} \texttt{bin\_bn}  \dots};

    \node [module, node distance=2cm, right of=dmod, fill=data1] (iid) {\texttt{iid}};
    \node [module, node distance=3cm, left of=gmod, text width=7em, fill=data1] (gmods) {\texttt{trilearn\_cta} \texttt{pcalg\_randdag} \dots};

    \node [block, text width=5em, above of=gmod, fill=data1] (seed) {\texttt{seed\_range} $k:m$};
    

    \node [cloud, below of=param, fill=data2] (pfile) {SEM, bn.fit};
    \node [cloud, below of=data, fill=data2, text width=3.7em] (dfile) {data file or folder};
    \node [cloud, below of=graph, fill=data2, ] (gfile) {graph file};
    
    \node [module, below of=pfile, text width=15em, node distance=2.5cm, fill=stl1] (stl) {\texttt{structure\_learning\_algorithms}};
    \node [module, right of=stl, text width=6em, node distance=5cm, fill=stl1] (stlmods) {\texttt{gobnilp} \texttt{tetrad\_fci} \texttt{pcalg\_pc} \dots};
    
    \node [cloud, below of=stl, fill=stl2] (ntest) {\#tests};
    \node [cloud, left of=ntest, fill=stl2] (adjmat) {graphs};
    \node [cloud, right of=ntest, fill=stl2] (time) {timings};
    \node [cloud, right of=time, node distance=3cm, fill=stl2] (gtraj) {MCMC trajectories};
    
    \node [module, below of=adjmat, node distance=2.5cm, text width=6em, fill=eval1] (bmarks) {\texttt{benchmarks}};
    \node [module, left of=graph, node distance=3.5cm, text width=9em, fill=data1] (gtplot) {\texttt{graph\_true\_plots} \texttt{graph\_true\_stats}};
    \node [module, left of=bmarks, node distance=3cm, text width=6em, fill=eval1] (gplot) {\texttt{graph\_plots}};
    \node [module, right of=data, node distance=3.5cm, text width=8em, fill=data1] (ggpairs) {\texttt{ggally\_ggpairs}};

    \node [module, below of=gtraj, node distance=3cm,text width=10em, fill=eval1] (mcmc) {\texttt{mcmc\_traj\_plots}  \texttt{mcmc\_autocorr\_plots} \texttt{mcmc\_heatmaps}};
    \node [cloud, below of=bmarks,text width=7em,node distance=3.5cm, fill=eval2] (output) {CSV summaries ROC plots graph types time boxplots SHD boxplots F1 boxplots \#tests boxplots \dots};
    
    \node [cloud, below of=gtplot,text width=4.5em, fill=data2] (gtfiles) {graphs heatmaps properties};
    
    \node [cloud, below of=mcmc, node distance=1.7cm,text width=7em, node distance=2.0cm, fill=eval2] (mcmcfiles) { trajectory plots autocorrelation heatmaps };
    \node [cloud, below of=gplot,text width=6.5em, fill=eval2] (gfiles) {graphs heatmaps graphviz.comp};
    \node [cloud, below of=ggpairs,text width=6em, fill=data2] (ggpairsfiles) {scatter plots};


 
    \path [uline] (iid) -- (dmod);
    \path [uline] (gmods) -- (gmod);
    \path [uline] (stlmods) -- (stl);
    \path [uline] (pmods) -- (pmod);
 
    \path [line] (param) -- (dmod);
    \path [altline,dashed] (gmod) -- (graph);
    \path [altline] (graph) -- (pmod);

    \path [altline,dashed] (pmod) -- (param);
    \path [altline,dashed] (dmod) -- (data);

    \path [altline, dashed] (gfile) -- (graph);
    \path [altline,dashed] (pfile) -- (param);
    \path [altline] (graph) -- (pfile);
    \path [altline,dashed] (dfile) -- (data);
    
    \path [altline] (data) -- (stl);
    
    \path [altline] (stl) -- (adjmat);
    \path [altline] (stl) -- (time);
    \path [altline] (stl) -- (ntest);
    \path [altline] (stl) -- (gtraj);
    \path [altline,draw=black] (graph) edge[bend right,draw=black] node [] {} (bmarks);
    \path [altline,draw=black] (data) -- (ggpairs);
    
    \path [altline] (bmarks) -- (output);

    \path [altline,draw=black] (adjmat) -- (gplot);
    \path [altline,draw=black!100] (graph) edge[bend right,draw=black!100] node [] {}  (gplot);

    \path [altline,draw=black] (adjmat) -- (bmarks);    
    \path [altline,draw=black] (time) -- (bmarks);
    \path [altline,draw=black!100] (ntest) -- (bmarks);
    \path [altline,draw=black] (gtraj) -- (mcmc);
    
    \path [altline,draw=black] (graph) -- (gtplot);
    \path [altline,draw=black] (gtplot) -- (gtfiles);
    \path [altline,draw=black] (mcmc) -- (mcmcfiles);
    \path [altline,draw=black] (gplot) -- (gfiles);
    \path [altline,draw=black] (ggpairs) -- (ggpairsfiles);
    
    
     \path [line] (seed) -- (gmod);
     \path [line] (seed) -- (pmod);
     \path [line] (seed) -- (dmod);
     \path [line] (seed) -- (stl);


\end{tikzpicture}
\caption{Flowchart for the \ttl architecture describing how the files and sections (sharp rectangles) of the \proglang{JSON}  configuration file are related to the modules (rounded rectangles with \code{monospace} font style). The different colours pink, blue, and purple indicate modules, files and sections related to data, structure learning, and evaluating results respectively. An arrow from a node A to another node B should be read as “B requires input from A”.  Thus, for any node, following the arrows in their opposite directions builds a path of the used modules or files. Dashed arrows indicate that one of the parents is required.}
\label{fig:flowchart}
\end{figure}

At the highest level, there are two main sections, \texttt{resources} and \texttt{benchmark\_setup}, described in Section~\ref{sec:resources} and \ref{sec:benchmark_setup}, respectively. 
The \texttt{resources} section contains separate subsections of the available \emph{modules} for specifying graphs (\texttt{graph}), parameters (\texttt{parameters}), data (\texttt{data}), and algorithms for structure learning (\texttt{structure\_learning\_algorithms}).
Each module, in turn, has a list, where each element is an object defining a parameter setting, identified by a unique  \texttt{id}.
The \texttt{benchmark\_setup} section specifies the data models (\texttt{data}) and evaluation methods (\texttt{evaluation}) a user wishes to consider for analysis. The performance evaluation of structure learning algorithms will depend not only on the chosen metric, but also on the particular graphical model of interest, and it may be measured on different spaces, e.g.\ CPDAGs, pattern graphs or skeletons, and these options are supported by \ttl.

The module objects used in \texttt{benchmark\_setup} are defined in \texttt{resources} and referenced by their corresponding \texttt{id}'s.
The output files of each module are saved systematically under the \emph{results/} directory based on the corresponding objects' values.
An important consequence of this design is that an algorithm run for a specific data set will never have to be rerun, instead, the stored output files are used for the different evaluation modules.

Figure~\ref{fig:flowchart} shows a flowchart describing how the files and sections relate to the modules in a \proglang{JSON} configuration file. 
The notation used on Figure~\ref{fig:flowchart} is described in more detail in Section~\ref{sec:json}.
Listings~\ref{json:pcvsdualpc} later shows an example of such a configuration file and is described in more detail in Section~\ref{sec:small}.
\subsection[Resources]{Resources section (\texttt{resources})}
\label{sec:resources}
Tables~\ref{tab:graph_mods}--\ref{tab:data_mods} shows the  available modules in  the  \code{graph}, \code{parameters}, and \code{data} sections.
For the modules that stem from existing packages, the names are conventionally prefixed by the name of the package on which they are based. 
In each of these sections, there are special modules, \code{fixed\_graph}, \code{fixed\_params}, and \code{fixed\_data}, enabling the user to provide graphs, parametrizations, and datasets using files. 
Note that, \ttl already contains several standard Bayesian network models, \emph{e.g.}, those available at the website of \pkg{bnlearn}, including the so-called \emph{Bayesian network repository}  \citep{friedman1997impact}.
Below is a brief description of the other available modules, for more details, we refer to Appendix \ref{app:modules} and the \ttl online documentation.


For the \code{graph} modules, the \code{bdgraph\_graphsim} module uses the \emph{graph.sim} function of the \pkg{BDgraph} package \citep{JSSv089i03} to sample various types of random undirected graphs.
The \code{pcalg\_randdag} module uses the \emph{randDAG} function of the \pkg{pcalg} package \citep{kalisch2012causal} to sample various types of random DAGs and undirected graphs.
The \code{trilearn\_bandmat} and \code{trilearn\_randbandmat} modules generate undirected decomposable graphs with a band-structured adjacency matrix of a given and random width, respectively.
The \code{trilearn\_cta} module samples decomposable graphs using the so-called \emph{Christmas tree algorithm} (CTA) \citep{cta}. 

\begin{table}[!ht]
    \centering
    \begin{tabular}{llll}
        \textbf{Method} & \textbf{Graph} & \textbf{Package} & \textbf{Module} \\ \hline
        Fixed graph &DAG, UG & - & \texttt{fixed\_graph}  \\ 
        randDAG & DAG, UG & \pkg{pcalg} & \texttt{pcalg\_randdag} \\ 
        graph.sim function & UG & \pkg{BDgraph} & \texttt{bdgraph\_graphsim} \\ 
        Bandmatrix & DG & \pkg{trilearn} & \texttt{trilearn\_bandmat} \\ 
        CTA & DG & \pkg{trilearn} & \texttt{trilearn\_cta} \\ 
        Random bandmatrix & DG & \pkg{trilearn} & \texttt{trilearn\_rand\_bandmat} \\ 
    \end{tabular}
    \caption{Graph modules.}
    \label{tab:graph_mods}
\end{table}

For the modules in the \code{parameters} section, the \pkg{bdgraph\_rgwish} module samples the precision matrix of an undirected Gaussian graphical model from the G-Wishart distribution \citep{dawid1993,Atay-Kayis01062005, lenkoski2013direct}.
The \code{binary\_bn} module samples the conditional probability tables of a Bayesian network with only binary variables.
The \code{sem\_params} module samples the weights of a structural equation model (SEM) Gaussian Bayesian network \eqref{eq:sem}.
The \code{trilearn\_hyper-dir} module samples the parameters of a multinomial decomposable graphical model from the Hyper-Dirichlet distribution \citep{dawid1993}.
The \code{trilearn\_intra-class} module specifies the covariance matrix of a Gaussian graph intra-class model.
\begin{table}[H]
    \centering
    \begin{tabular}{llll}

        \textbf{Method} & \textbf{Graph}  & \textbf{Package} & \textbf{Module} \\ \hline
        Fixed parameters &  DAG & - & \texttt{fixed\_params} \\ 
        SEM parameters & DAG  & \pkg{Benchpress} & \texttt{sem\_params} \\ 
        Binary BN & DAG  & \pkg{Benchpress} & \texttt{binary\_bn}\\ 
        Hyper-Dirichlet & DG  & \pkg{trilearn} & \texttt{trilearn\_hyper-dir} \\ 
        Graph intra-class & UG  & \pkg{trilearn} & \texttt{trilearn\_intra-class} \\ 
        G-Wishart & UG  & \pkg{BDgraph} & \texttt{bdgraph\_rgwish} \\ 
    \end{tabular}
    \caption{Parameterisation modules.}
    \label{tab:params_mods}
\end{table}

For the modules in the \code{data} section, the \code{iid} module is a generic module to draw independent identically distributed (i.i.d.) samples from any of the models built by proper combinations of the \texttt{graph} and \texttt{parameters} modules.
The \code{gcastle\_iidsim} module uses the \emph{IIDSimulations} function of the \pkg{gCastle} package to draw i.i.d.\ samples from different types of SEM models.





\begin{table}[!ht]
    \centering
    \begin{tabular}{llll}
        \textbf{Method} & \textbf{Graph} & \textbf{Package} & \textbf{Module} \\ \hline
        Fixed dataset(s) & all & -& \texttt{fixed\_data}  \\ 
        gCastle i.i.d. (SEM) & DAG & \pkg{gCastle} & \texttt{gcastle\_iidsim} \\
        I.i.d. data & all & various ~ &\texttt{iid}  \\ 
    \end{tabular}
    \caption{Data modules.}
    \label{tab:data_mods}
\end{table}


The available modules in the  \code{structure\_learning\_algorithms} section are listed in Table~\ref{tab:algorithms}.
For the algorithms demonstrated in the simulation studies in Section~\ref{sec:case_study}, we give a short overview in Appendix~\ref{app:modules}.


\subsection[Benchmark setup]{Benchmark setup (\texttt{benchmark\_setup})}\label{sec:benchmark_setup}
\subsubsection[Data setup section]{Data setup section (\code{data})}

The \texttt{data} section consists of a list of objects,  each containing the \texttt{id} of a graph (\texttt{graph\_id}), parameters (\texttt{parameters\_id}), data (\texttt{data\_id}) module object, and a tuple (\texttt{seed\_range}), specifying the range of the random seeds used in the modules.
For each seed number \(i\) in the specified range, a graph \(\graph_i\) is obtained as specified according to \(\texttt{graph\_id}\).
Given \(\graph_{i}\), the parameters in the model $\Theta_i$ are obtained according to \texttt{parameters\_id}.
A data matrix, $\rma_i=(\rva_{1:\p}^j)_{j=1}^n$, is then sampled from $(\graph_i,\Theta_i)$ according to the model and \texttt{data\_id}.
When combining modules from the different sections, it is important to ensure that $(\graph_i,\Theta_i)$ is a compatible pair, \emph{e.g.}, if $\Theta_i$ represents the parameters of Bayesian network then $\graph_i$ is required to be a DAG.

\begin{center}
\begin{table}[H]
\begin{center}
\begin{tabular}{r|lll}
\textbf{} &  \(\graph\) & \(\param\) & \(\rma\) \\\hline 
I         & -              & -                   & Fixed         \\ 
II        & Fixed          & -                   & Fixed         \\
III       & Fixed          & Fixed               & Generated     \\
IV        & Fixed          & Generated           & Generated     \\
V         & Generated      & Generated           & Generated    
\end{tabular}
\caption{Data scenarios.}
\label{tab:datascen}
\end{center}
\end{table}
\end{center}

By using the modules for fixed files, we can build models with increased flexibility.
The different data sources provided by \ttl can be summarised in five scenarios shown in Table~\ref{tab:datascen}.
Scenario I is the typical scenario for data analysts, where users provide one or more datasets by hand.  
Scenario II is similar to Scenario I, where the difference is that in addition, the user provides the true graph underlying the data.
This situation arises \emph{e.g.}, when replicating a simulation study from the literature, where both the true graph and the dataset are given.
Scenarios III-V are pure benchmarking scenarios, where either all of the graphs, parameters, and data are generated (V), or the graphs (III, IV) and possibly parameters (III) are specified by the user.

Note that, when a filename is used as \texttt{data\_id}, since the data is fixed, \texttt{seed\_range} should be set to \vals{null}.
However, depending on the evaluation module used, the \texttt{graph\_id} could be either \vals{null} or the filename of the true graph underlying the data.
For example, the \texttt{benchmarks} module requires that \texttt{graph\_id} set to a proper \texttt{id} for a graph module object, however, it is not a requirement for the \texttt{adjmat\_plots} module.
When the value of \texttt{data\_id} is a directory, the other fields should be set to \vals{null}.

\subsubsection[Evaluation section]{Evaluation section  (\code{evaluation})} 
\label{sec:evaluation}

Table~\ref{tab:eval_mods} shows the available modules in the \texttt{evaluation} section.
The figures for each module are saved systematically in \emph{results/} based on the parameterisation of the module used and copied to \emph{results/output/} for easy reference.
Further details are provided in Appendix \ref{app:modules}.
\begin{table}[ht]
    \centering
    \begin{tabular}{ll}
        \textbf{Evaluation} & \textbf{Module} \\ \hline
        Benchmarks & \texttt{benchmarks} \\ 
        Pairs plot & \texttt{ggally\_ggpairs} \\
        Graph plots & \texttt{graph\_plots} \\ 
        True graph plots & \texttt{graph\_true\_plots} \\
        True graph properties & \texttt{graph\_true\_stats} \\ 
        MCMC auto-correlation & \texttt{mcmc\_autocorr\_plots} \\ 
        MCMC mean graphs & \texttt{mcmc\_heatmaps} \\ 
        MCMC trajectories & \texttt{mcmc\_traj\_plots} \\ 
    \end{tabular}
    \caption{Evaluation modules.}
    \label{tab:eval_mods}
\end{table}

The \code{benchmarks} module provides timings and comparisons using standard benchmarking metrics.
The results are summarised in \emph{e.g.}\ \emph{box plots} and \emph{receiver operating characteristic} (ROC)-type curves. 
In addition to the figures produced, the raw data is saved in \emph{comma separated values} (CSV) files which could be analysed externally.
The \code{ggally\_ggpairs} module uses the \emph{ggpairs} function of the \pkg{GGally} package \citep{emerson2013generalized} to visualise the data in pairwise scatter plots.
The \texttt{graph\_plots} module plots and saves the estimated graphs and adjacency matrices. If the true graph is available (scenarios II-V in Table~\ref{tab:datascen}) it also compares the true to the estimated graphs using the \emph{graphviz.compare} function from the \pkg{bnlearn} package \citep{JSSv035i03}.
The \texttt{graph\_true\_plots} module plots the true underlying graphs and corresponding adjacency matrices.
The \texttt{graph\_true\_stats} module computes and plots properties of the true underlying graphs.
For the MCMC algorithms, the following three special evaluation modules are available.
The \texttt{mcmc\_heatmaps} module estimates posterior edge probabilities and plots them in heatmaps using the \pkg{seaborn} package \citep{Waskom2021}.
The \texttt{mcmc\_traj\_plots} module plots the value of a given functional for the graphs in an MCMC trajectory. 
The currently supported functionals are the number of edges for the graphs and the scores.
The \texttt{mcmc\_autocorr\_plots} module plots the auto-correlation of a functional of the graphs in a MCMC trajectory.

\section{Benchpress in practice} \label{sec:case_study}
We demonstrate the functionalities of \ttl with seven examples for benchmarking structure learning algorithms for Bayesian networks.
We consider the pure benchmarking setting for both binary and continuous data types as well as a fixed data scenario (Section~\ref{sec:sachs}). 
For performance evaluation and visualisation, we use the modules \texttt{benchmarks}, \texttt{graph\_true\_stats}, \texttt{ggally\_ggpairs}, and \texttt{graph\_plots}.
The title rows of each sub-figure describe the settings for the graph, parameters, data modules, and the number of datasets used.

We use the \texttt{benchmarks} module to show scatter plots of the individual results in terms of ROC type curves, where median $FP/P$ is plotted against the median $TP/P$ and estimated 5\%-95\% percentiles in ROC type figures (see Appendix \ref{appendix:metrics} for definitions). 
Since not all methods support the same test procedures or scores, we have chosen the default settings for some of the algorithms.
The parameter values are selected so as to indicate how the performance changes.
However, some curves are shorter than would be desired due to range limits for the parameters in the programs used.

For \textbf{binary data} we use the \emph{Bayesian Dirichlet likelihood-equivalence} (BDe) score \citep{heckerman1995learning} in the score-based methods while varying the equivalent sample size parameter \citep{silander2012sensitivity}. 
The benchmarking setup considers the following score-based algorithms: BOSS (\texttt{boss-bdeu}), GRaSP (\texttt{grasp-bdeu}), 
HC (\texttt{hc-bde}), 
Tabu (\texttt{tabu-bde}), ASOBS (\texttt{asobs-bdeu}), GOBNILP (\texttt{gobnilp-bde}), Iterative search  (\texttt{itsearch-bde}), and order MCMC with the end space of \texttt{itsearch-bde} as start space  (\texttt{omcmc-bde}).
Furthermore, we include the constraint-based PC algorithm with the \(G^2\)-test (\texttt{pc-binCItest}) and the hybrid method MMHC (\texttt{mmhc-bde-mi})
with the \emph{mutual information} test and the BDe score. 
The varying parameter to draw ROC-type curves for all constraint-based methods is the significance level for the independence tests.

For the \textbf{continuous datasets}, we use two different scores in the score-based methods. The benchmarking setup in this case includes  HC (\texttt{hc-bge}), Tabu (\texttt{tabu-bge}), GOBNILP (\texttt{gobnilp-bge}),
Iterative search (\texttt{itsearch-bge}), and order MCMC using the end space of \texttt{itsearch-bge} as start space (\texttt{omcmc-bge}) with the \emph{Bayesian Gaussian likelihood-equivalence} (BGe) score \citep{geiger1994learning,geiger2002parameter, kuipers2014addendum}, and varying levels of the prior parameter.
Furthermore we include the FGES (\texttt{fges-sem-bic}), BOSS (\texttt{boss-sem-bic}), and GRaSP (\texttt{grasp-sem-bic}) with the \emph{structural equation model Bayesian information criterion} (SEM-BIC) score \citep{glymour1986causal}, and varying levels of the penalty discount.
We also include the NO TEARS algorithm (\texttt{notears-l2}) with the \(L_2\) penalty.
For constraint-based methods we include the PC algorithm (\texttt{pc-gaussCItest}), dual PC algorithm (\texttt{dualpc}), and MMHC (\texttt{mmhc-bge-zf}) 
with the \emph{Fisher’s z-transformation} test of partial correlations \citep{edwards2012introduction} with varying significance levels.

\subsection{Comparative study between the PC and the dual PC algorithm}\label{sec:small}

We start with a simple demonstration comparing just two algorithms side-by-side and a small-scale simulation study from data scenario V with the PC and the dual PC algorithms.
We consider data from 10 random Bayesian network models \(\{(\graph_i,\param_i)\}_{i=1}^{10}\), where each graph \(\graph_i\) has \(\p=80\) nodes and is sampled using the \randdag\, module. 
The parameters \(\param_i\) are sampled from the random linear Gaussian SEM using the \texttt{sem\_params}\, module (Appendix~\ref{app:params}) with \(a=0.25, b=1, \mu=0\) and \(\sigma=1\).
We draw a standardised dataset \(\rma_i\) of size \(n=300\) from each model using the \iid\, module. 
Listing~\ref{json:pcvsdualpc} shows the content of the \proglang{JSON} configuration file (\emph{paper\_pc\_vs\_dualpc.json}) for this study.

\begin{listing}
    \begin{minted}[frame=single,
                   framesep=3mm,
                   fontsize=\tiny,
                   linenos=true,
                   xleftmargin=21pt,
                   tabsize=4]{js}
{
    "benchmark_setup": {
        "data": [
            {
                "graph_id": "avneighs_p20",
                "parameters_id": "SEM",
                "data_id": "standardized",
                "seed_range": [1, 10]
            }
        ],
        "evaluation": {
            "benchmarks": {
                "filename_prefix": "paper_pc_vs_dualpc/",
                "show_seed": false,
                "errorbar": true,
                "errorbarh": false,
                "scatter": true,
                "path": true,
                "text": false,
                "ids": ["pc-gaussCItest", "dualpc"]
            },
            "graph_true_plots": true,
            "graph_true_stats": true,
            "ggally_ggpairs": false,
            "graph_plots": ["pc-gaussCItest", "dualpc"],
            "mcmc_traj_plots": [],
            "mcmc_heatmaps": [],
            "mcmc_autocorr_plots": []
        }
    },
    "resources": {
        "data": {
            "iid": [
                {
                    "id": "standardized",
                    "standardized": true,
                    "n": 300
                }
            ]
        },
        "graph": {
            "pcalg_randdag": [
                {
                    "id": "avneigs4_p20",
                    "max_parents": 5,
                    "n": 20,
                    "d": 4,
                    "par1": null,
                    "par2": null,
                    "method": "er",
                    "DAG": true
                }
            ]
        },
        "parameters": {
            "sem_params": [
                {
                    "id": "SEM",
                    "min": 0.25,
                    "max": 1
                }
            ]
        },
        "structure_learning_algorithms": {
            "dualpc": [
                {
                    "id": "dualpc",
                    "alpha": [0.001, 0.05, 0.1],
                    "skeleton": false,
                    "pattern_graph": false,
                    "max_ord": null,
                    "timeout": null
                }
            ],
            "pcalg_pc": [
                {
                    "id": "pc-gaussCItest",
                    "alpha": [0.001, 0.05, 0.1],
                    "NAdelete": true,
                    "mmax": "Inf",
                    "u2pd": "relaxed",
                    "skelmethod": "stable",
                    "conservative": false,
                    "majrule": false,
                    "solveconfl": false,
                    "numCores": 1,
                    "verbose": false,
                    "indepTest": "gaussCItest",
                    "timeout": null
                }
            ]
        }
    }
}
 \end{minted}
\caption{The \proglang{JSON} configuration file \emph{paper\_pc\_vs\_dualpc.json} to define a small comparison between the PC and dual PC algorithms.} 
\label{json:pcvsdualpc}
\end{listing}

\begin{figure}[!ht]

    \begin{subfigure}[b]{.45\textwidth}
      \centering
      \includegraphics[width=1.0\linewidth]{{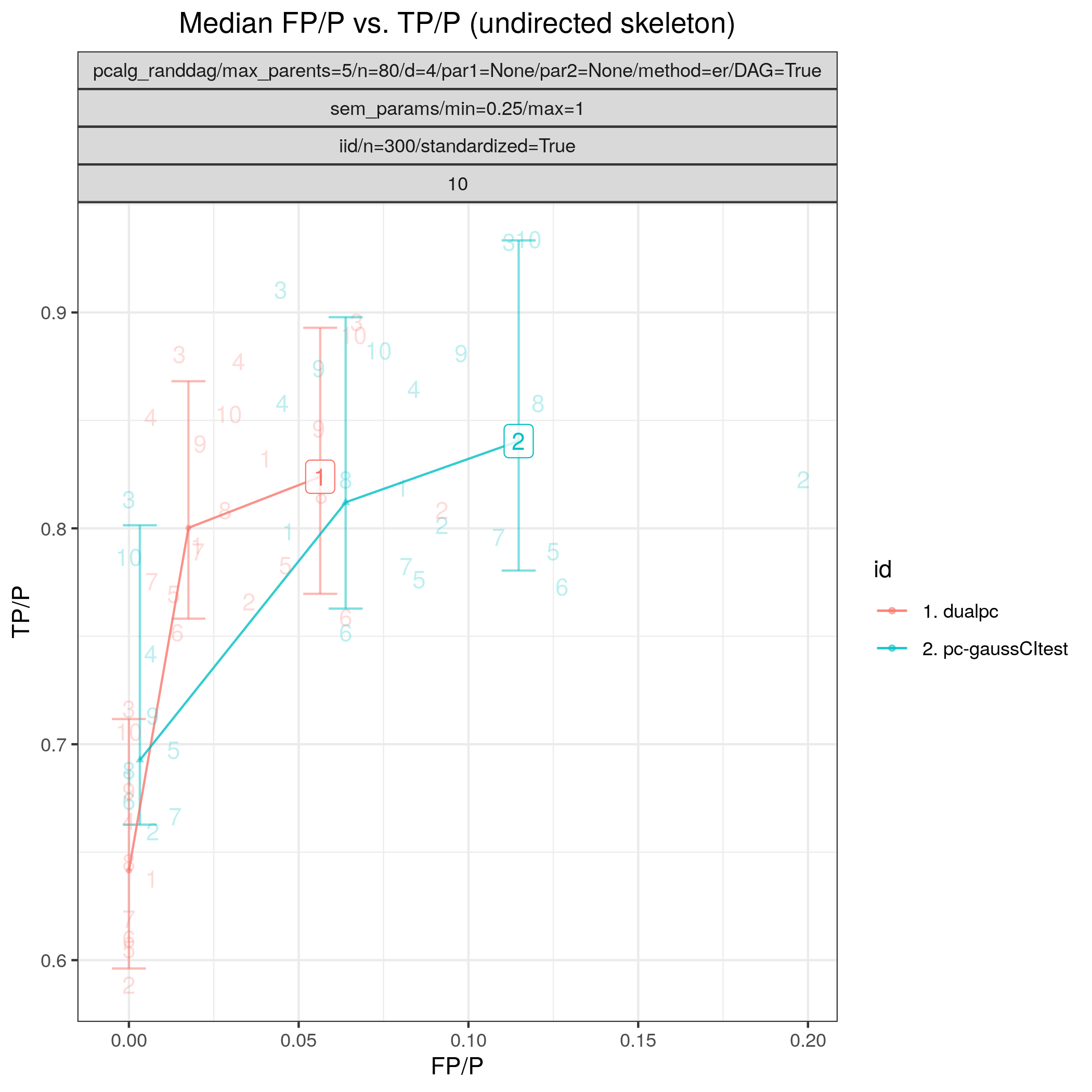}}  
      \caption{}
        \label{fig:small_roc}
    \end{subfigure}
        ~
    \begin{subfigure}[b]{.45\textwidth}
      \centering
      \includegraphics[width=1.0\linewidth]{{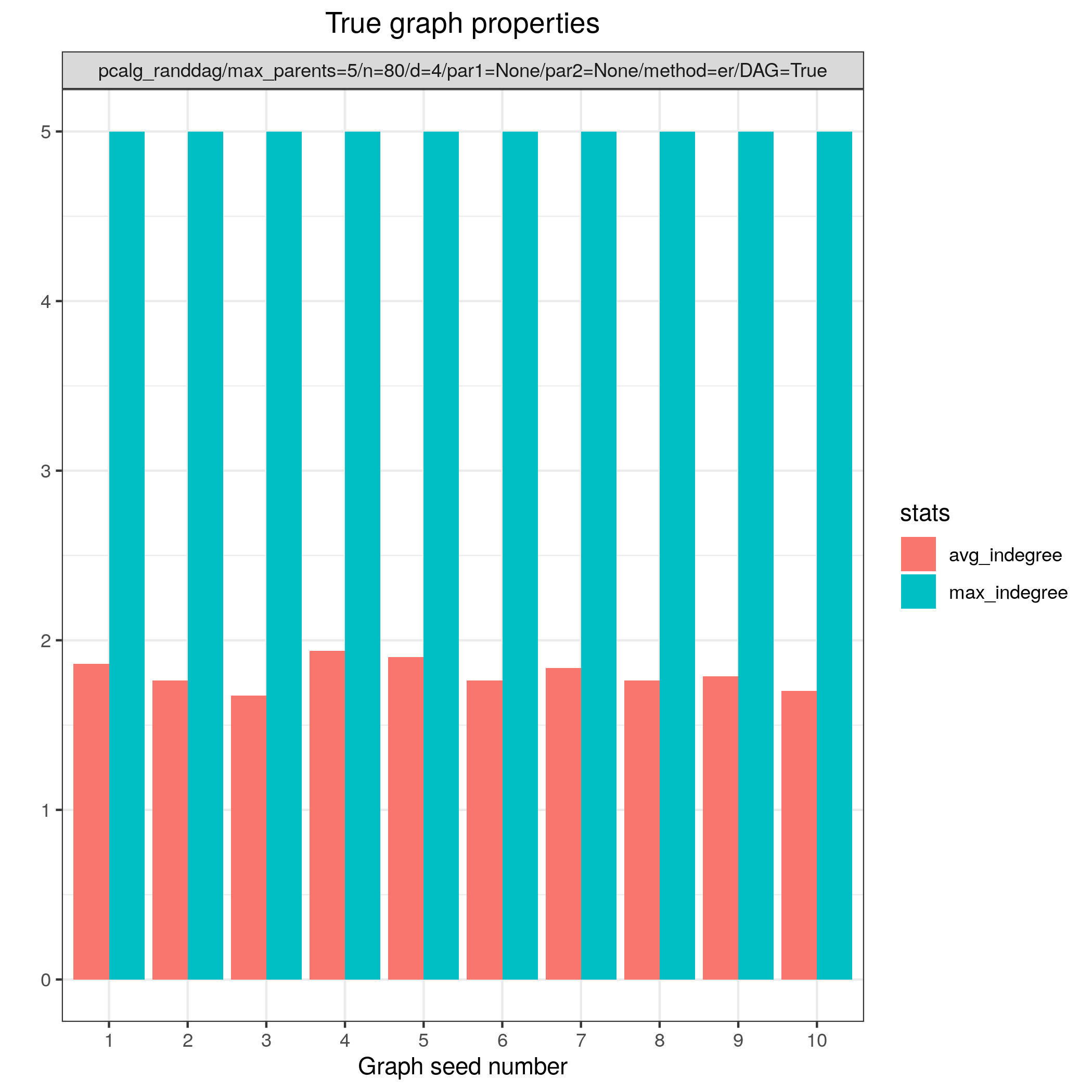}}  
      \caption{}
        \label{fig:small_density}
    \end{subfigure}
    
    \begin{subfigure}[b]{.45\textwidth}
      \centering
      \includegraphics[width=1.0\linewidth]{{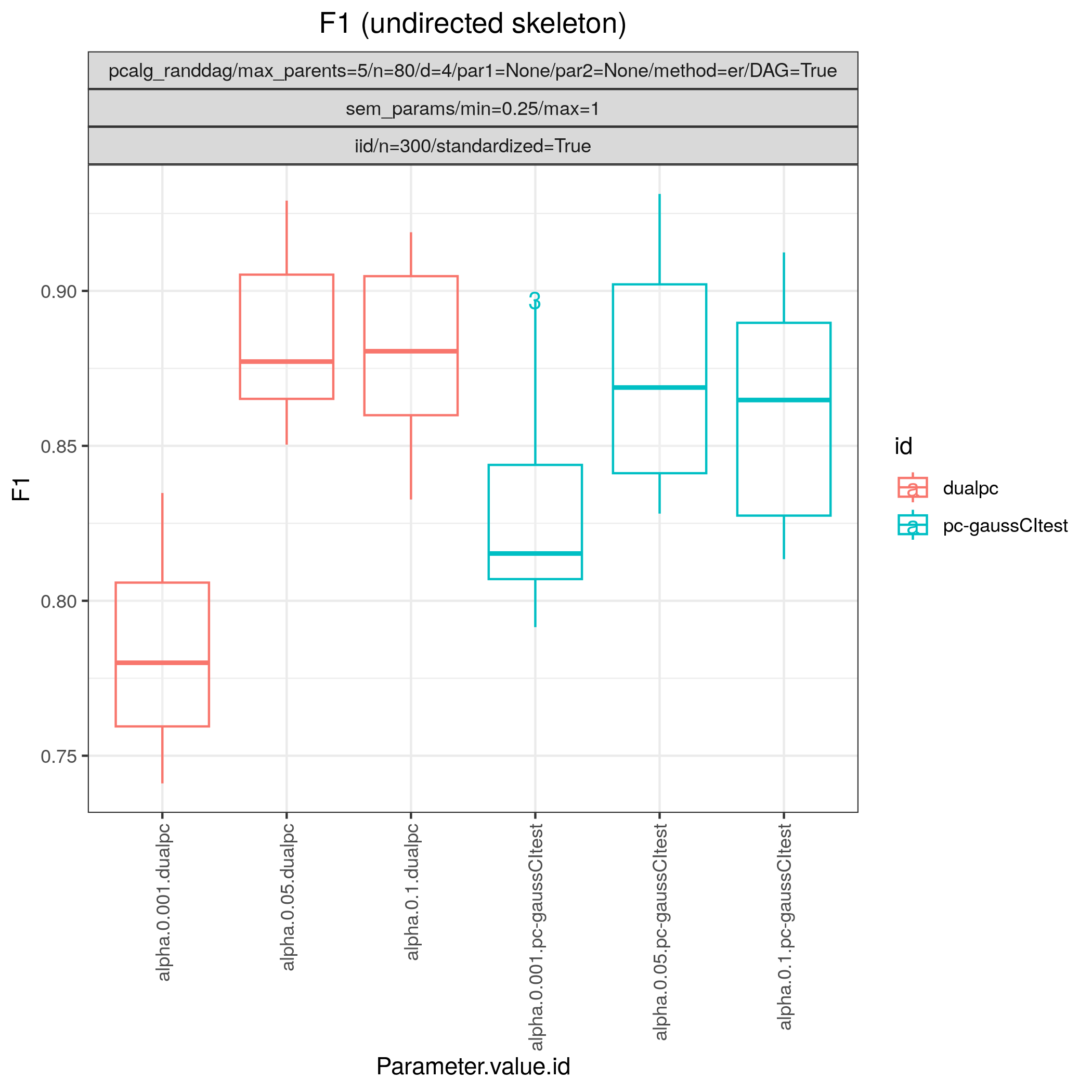}}  
      \caption{}
        \label{fig:small_f1}
    \end{subfigure}
    ~
    \begin{subfigure}[b]{.45\textwidth}
      \centering
      \includegraphics[width=1.0\linewidth]{{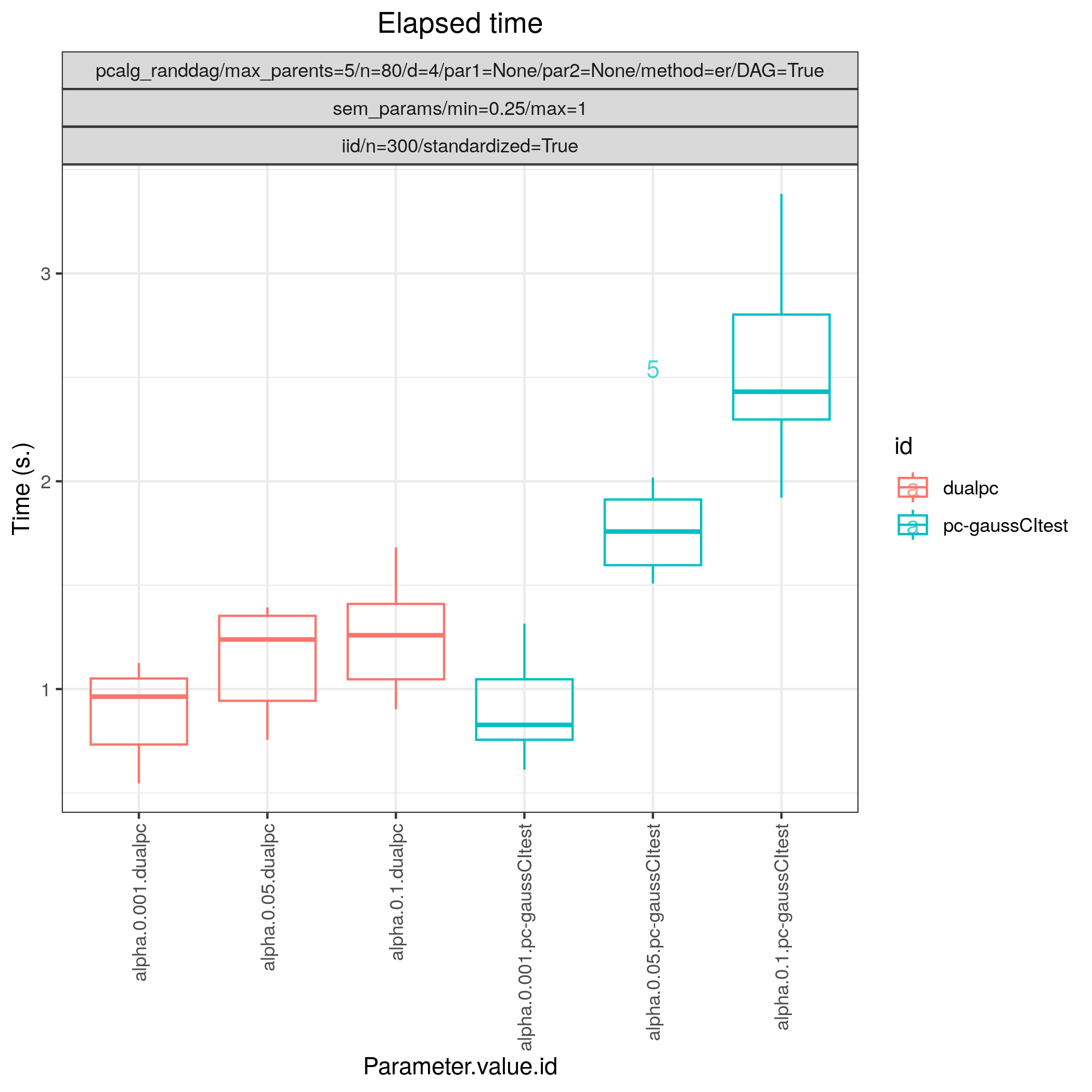}}  
      \caption{}
        \label{fig:small_time}
    \end{subfigure}
    \caption{
    This figure shows results for a comparative study between the PC \texttt{pcalg\_pc} and the dual PC \texttt{dualpc} algorithm.
    In (\ref{fig:small_roc}) $FP/P$ (and median $FP/P$) is plotted against $TP/P$ (and median $TP/P$) in a ROC-type figure based on the \texttt{graph}, \texttt{parameters}, and \texttt{data} modules used. The parameter values are selected so as to indicate how the performance changes.
The upper and lower bars in the plots show the estimated 5\% and 95\% quantiles of the $TP/P$ for each module setting.
$F_1$ scores based on the graph skeletons and computational times are visualised by box-plots in (\ref{fig:small_f1}) and (\ref{fig:small_time}), respectively, where outliers are indicated by the corresponding seed number. Graph properties calculated from the underlying true graphs are shown in (\ref{fig:small_density})  and provided by the \texttt{graph\_true\_stats} module. 
All the plots are based on graphs and datasets corresponding to 10 different seed numbers.}
    \label{fig:small}
\end{figure}

Figure~\ref{fig:small} shows results from the \texttt{benchmarks} and the \texttt{graph\_true\_stats} module, where we have focused on the undirected skeleton for evaluations since this is the part where the algorithms mainly differ.
More specifically, from Figure~\ref{fig:small_roc}, showing the $FP/P$ and $TP/P$, we see that the dual PC has superior performance for significance levels \(\alpha=0.05,0.01\).
Apart from the curves, the numbers in the plot indicate the seed number of the underlying dataset and models for each run.
We note that the model with seed number 3 seems to give good results for both algorithms and looking into Figure~\ref{fig:small_density}, we note that the graph with seed number 3 corresponds to the one with the lowest graph density (\(|\graphedgeset| / |\graphnodeset|\)).
The box plots from Figure~\ref{fig:small_time} show the computational times for the two algorithms, where the outliers are labelled by the model seed numbers.
We note \emph{e.g.},\ that seed number 1 gave longer computational time for the standard PC algorithm and from Figure~\ref{fig:small_density} we find that the graph with seed number 1 has relatively high graph density.
The conclusion of the $F_1$ score plot in Figure~\ref{fig:small_f1} is in line with the $FP/P$ vs. $TP/P$ results from Figure~\ref{fig:small_roc}.

\subsection{Biological dataset with fixed DAG}\label{sec:sachs}
Next, we consider the data from \cite{sachs2005causal} containing cytometry measurements of 11 phosphorylated proteins and phospholipids, which has become common in this field since the true underlying graph is regarded as known, and as such we include a wider range of algorithms to compare.
The dataset consists of 7644 measurements in total, from nine different perturbation conditions, each defining a unique intervention scheme.
\cite{sachs2005causal} removed any data points that fell more than three standard deviations from the mean. 
The data were then discretized on three levels. 
They also use bootstrapping methodologies and handle the interventional dataset to determine the causal directions of edges. 

\begin{figure}[!ht]
  \centering
  \includegraphics[width=0.7\linewidth]{{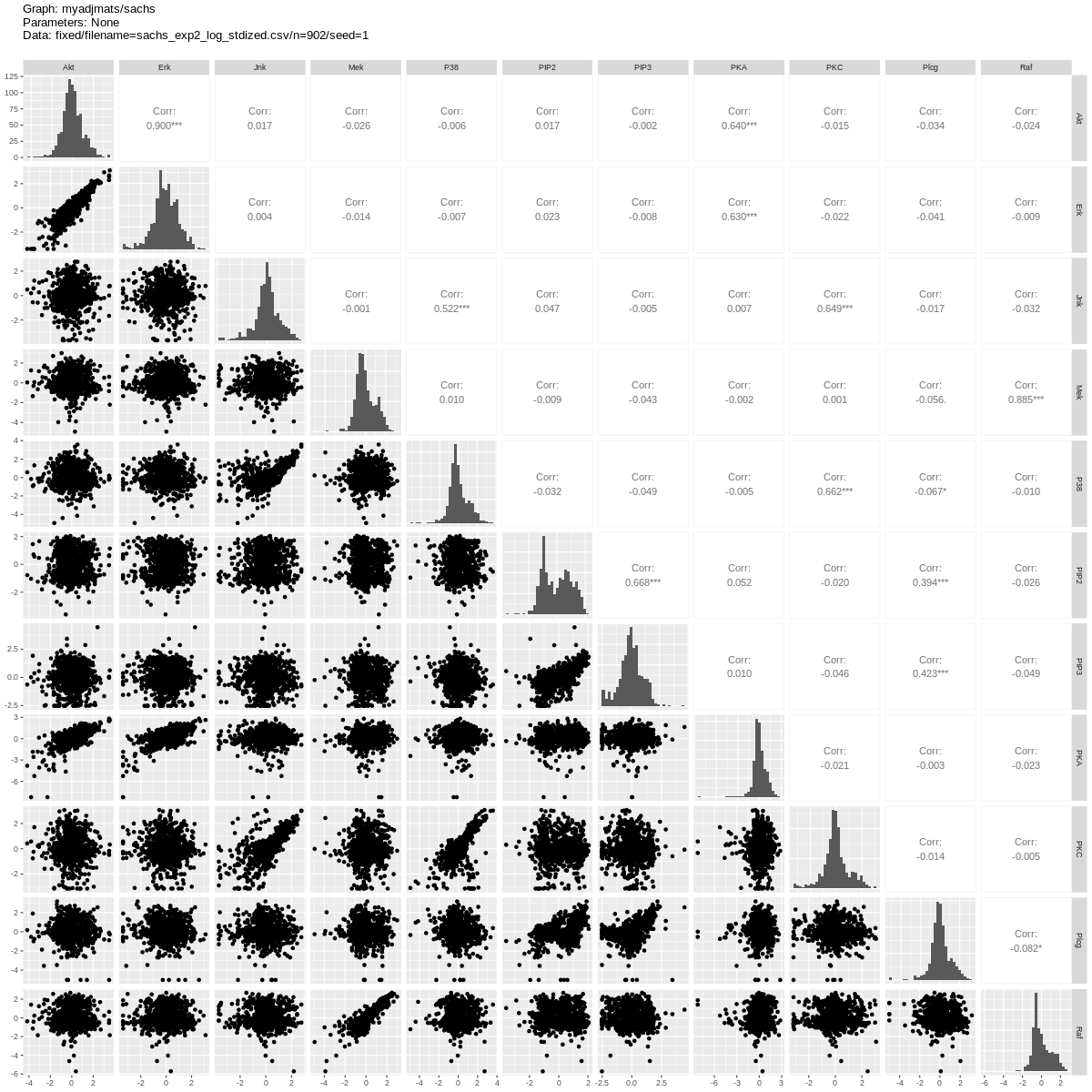}}  
\caption{Scatter plots of the logged and standardized data from the second experiment in \cite{sachs2005causal} by the \texttt{ggally\_ggpairs} module.}
\label{fig:sachs_pairs}
\end{figure}

However, since the purpose here is to benchmark algorithms suited for observational data, we consider only the first two interventions, referred to as \emph{(anti-CD3/CD28)} and \emph{(anti-CD3/CD28 + ICAM-2)} as only these interventions are expected to be independent of the nodes in the network and just activate the T-cells generally. 
\ttl also provide support for interventional data \citep{hauser2012characterization,NIPS2017_275d7fb2, kuipers2022interventional} through the \texttt{pcalg\_gies} module \citep{hauser2012characterization}, though we focus on the observational case here.
We show results for the (logged and standardized version of) the second dataset (\emph{anti-CD3/CD28 + ICAM-2}) with 902 observations since the graphs estimated from this dataset were in general closer to the gold standard network. The data are visualised in Figure~\ref{fig:sachs_pairs} with independent and pairwise scatter plots using the \texttt{ggally\_ggpairs} module.

Listing~\ref{json:sachs} shows the object in the \texttt{data} section of the config file defining the data setup. 
This setup falls into data Scenario II of Table~\ref{tab:datascen} since the \texttt{graph\_id} is set to the filename of the graph.
For Scenario I, when the underlying graph is unknown, \texttt{graph\_id} would be set to \vals{null}.
The full \proglang{JSON} specification for this study is found in \emph{paper\_sachs.json}.

\begin{listing}
    \begin{minted}[frame=single,
                   framesep=3mm,
                   linenos=true,
                   xleftmargin=21pt,
                   tabsize=4]{js}

{
    "graph_id": "sachs.csv",
    "parameters_id": null,
    "data_id": "2005_sachs_2_cd3cd28icam2_log_std.csv",
    "seed_range": null
}
        
\end{minted}
\caption{Scenario II \texttt{data} object for the transformed second dataset from \cite{sachs2005causal}.} 
\label{json:sachs}
\end{listing}

\begin{figure}[!ht]
 \centering
 \begin{subfigure}[t]{.45\textwidth}
  \centering
  \includegraphics[width=1.0\linewidth]{{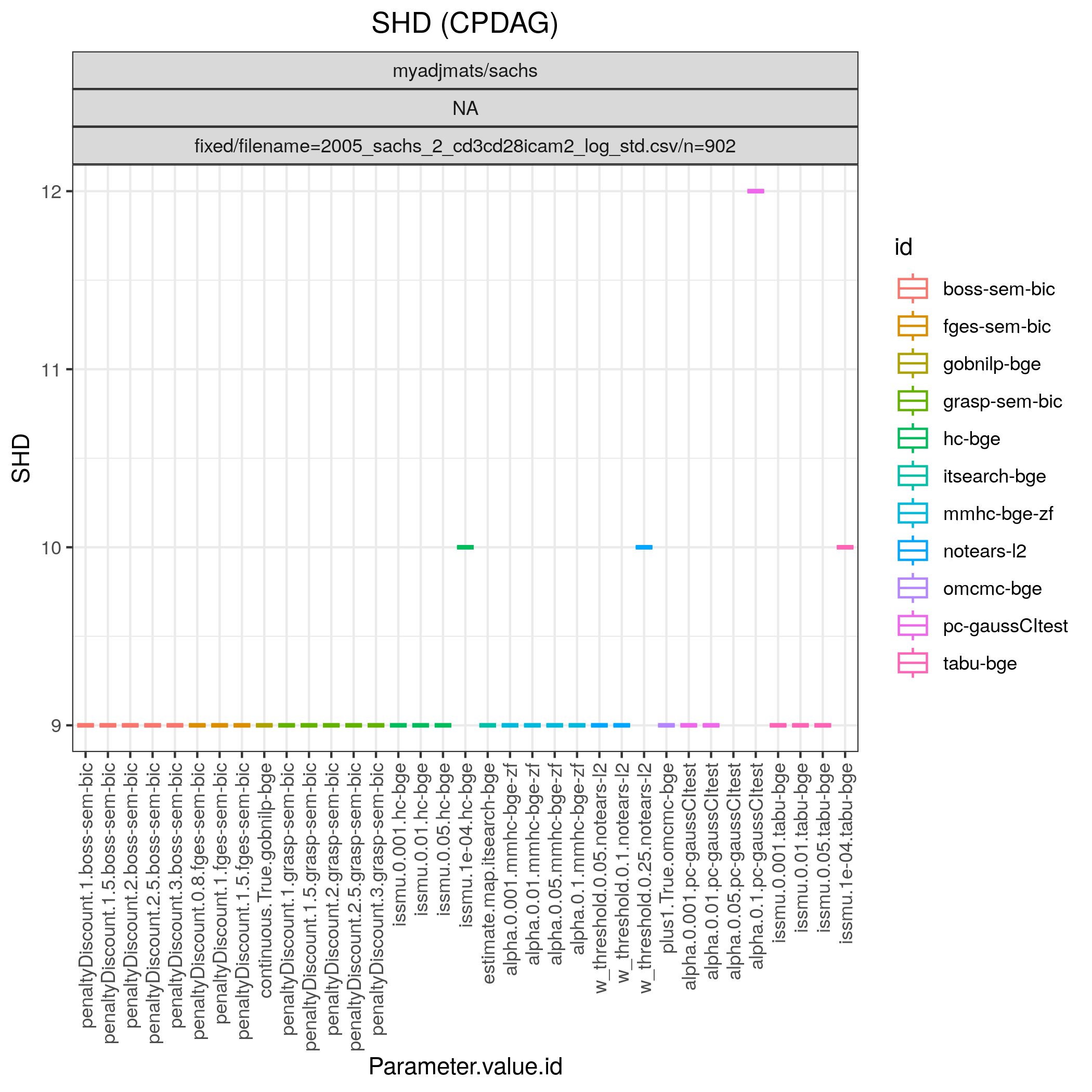}}  
  \caption{}
    \label{fig:sachs_shd}
\end{subfigure}
 \begin{subfigure}[t]{.45\textwidth}
  \centering
  \includegraphics[width=1.0\linewidth]{{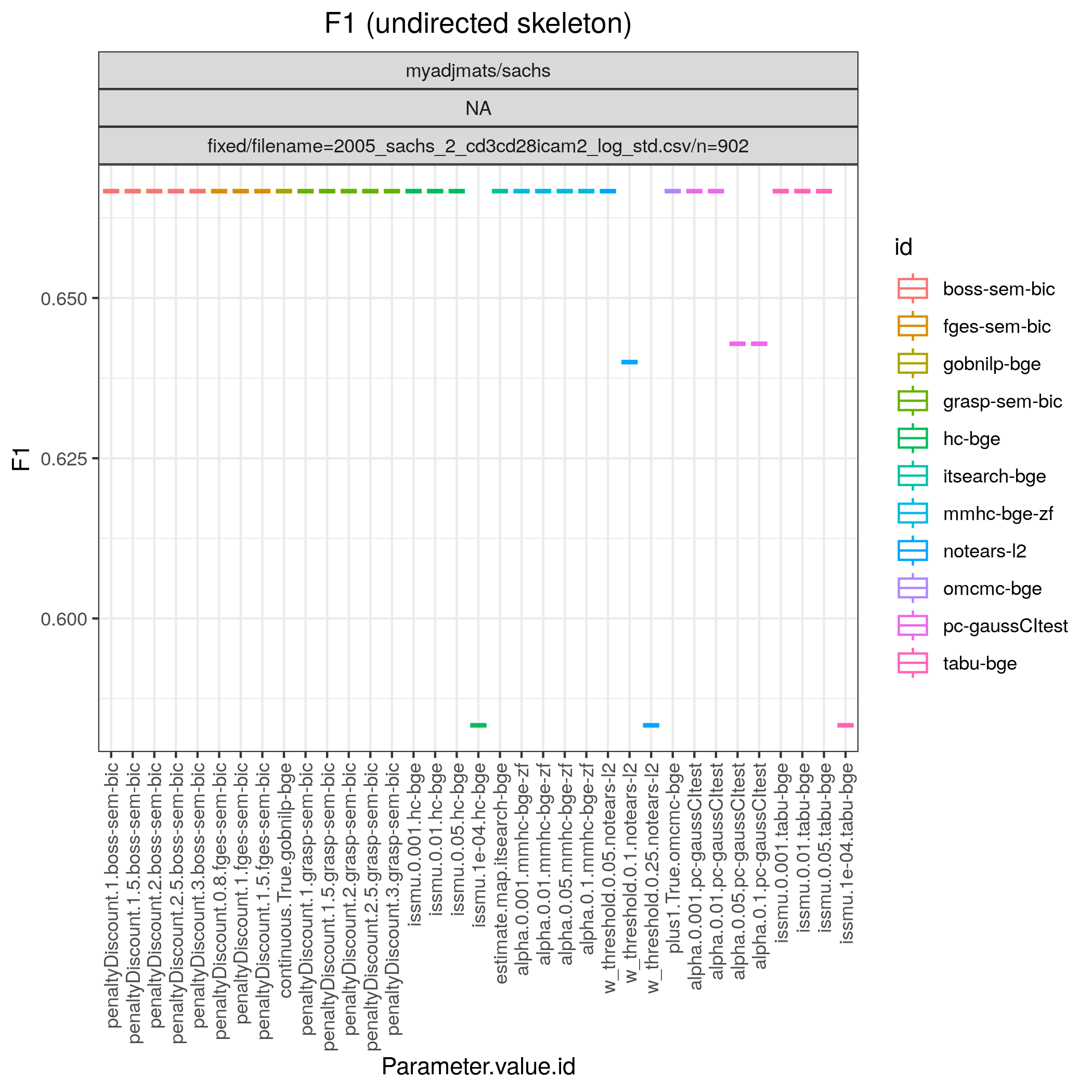}}  
  \caption{}
    \label{fig:sachs_f1}
\end{subfigure}
\caption{SHD (\ref{fig:sachs_shd}) based on the CPDAG and $F_1$ score (\ref{fig:sachs_f1}) based on the skeleton produced by the \texttt{benchmarks} module.}
\end{figure}

\begin{figure}[tb]

\begin{subfigure}[t]{.55\textwidth}
  \centering
  \includegraphics[width=1.0\linewidth]{{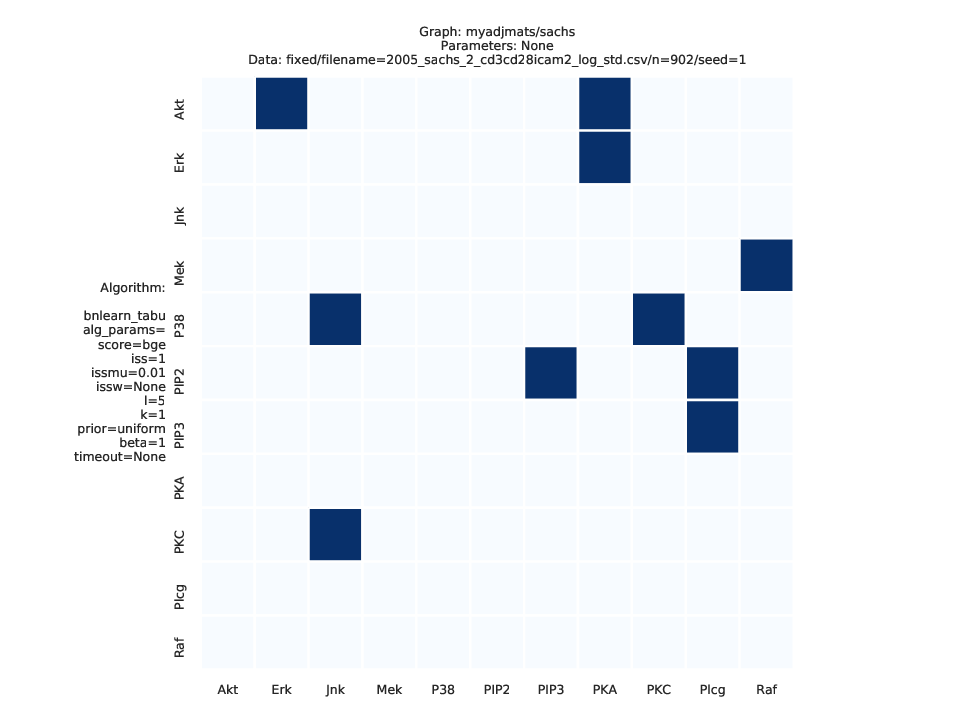}}  
  \caption{}
    \label{fig:sachs_adjmat}
\end{subfigure}
\begin{subfigure}[t]{.35\textwidth}
  \centering
  \includegraphics[width=1.0\linewidth, page=2]{{figures/sachs/graph_29.png}}  
  \caption{}
    \label{fig:sachs_graphest}
\end{subfigure}

\caption{Adjacency matrix (\ref{fig:sachs_adjmat}) and graph (\ref{fig:sachs_graphest}) plots produced by the \texttt{graph\_plots} module for the Tabu estimate. }
\label{fig:graph_plots}
\end{figure}

\begin{figure}[!t]
\centering
\begin{subfigure}[t]{.35\textwidth}
  \centering
    \includegraphics[width=1.0\linewidth, page=1]{{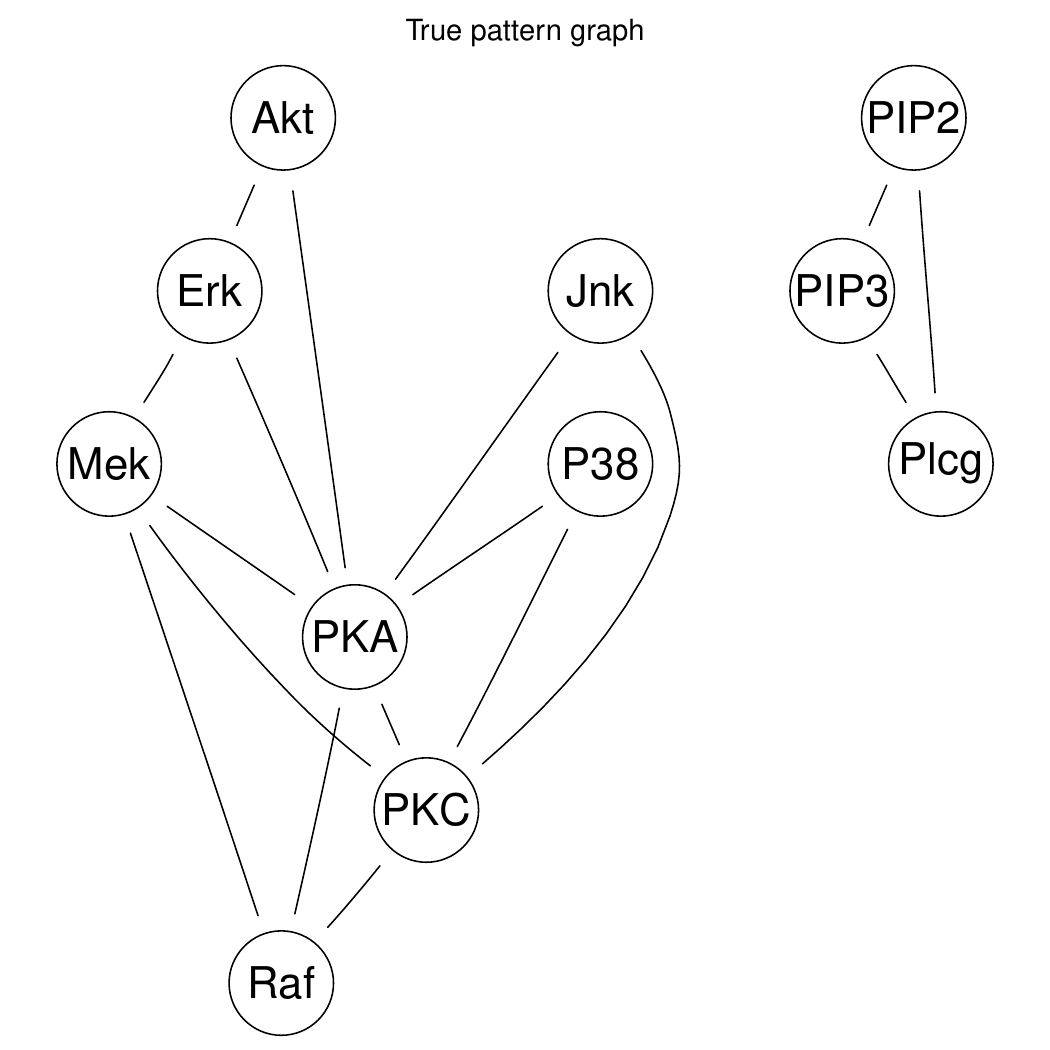}}  
  \caption{}
    \label{fig:sachs_compare1}
\end{subfigure} \hspace{0.1\textwidth}
\begin{subfigure}[t]{.35\textwidth}
  \centering
  \includegraphics[width=1.0\linewidth, page=2]{{figures/sachs/compare_19.pdf}}  
  \caption{}
    \label{fig:sachs_compare2}
\end{subfigure}

\caption{True pattern graph  (\ref{fig:sachs_compare1}) along with a comparison to the estimated pattern graph (\ref{fig:sachs_compare2}), using the \texttt{graph\_plots} module for the Tabu estimate shown in  Figure~\ref{fig:graph_plots}), with parameters as in Figure~\ref{fig:sachs_adjmat}).}
\label{fig:sachs_compare}
\end{figure}

Figure~\ref{fig:sachs_shd} shows SHD based on the CPDAG and $F_1$ score based on the undirected skeleton from 11 algorithms with different parametrisations, produced by the \texttt{benchmarks} module.
From this figure we can directly conclude that all algorithms have a parametrisation that gives the minimal SHD of 9 and maximal $F_1$ score of 0.67.
Figure~\ref{fig:sachs_adjmat} and Figure~\ref{fig:sachs_graphest} show the adjacency matrix and DAG plots, respectively, produced by the \texttt{graph\_plots} module of the DAG estimated by the \texttt{bnlearn\_tabu} module.
Figure~\ref{fig:sachs_compare} shows the pattern graph of both the true (Figure~\ref{fig:sachs_compare1}) and a DAG (Figure~\ref{fig:sachs_compare2}) estimated by the \texttt{bnlearn\_tabu} module, where the black edges are correct in both subfigures. 
The missing and incorrect edges are coloured in blue and red respectively in Figure~\ref{fig:sachs_compare2}.

\subsection{Linear Gaussian SEM with random weights and random DAGs (small scale study)}\label{sec:reproducible}
In the present study we consider a broader simulation over 11 algorithms in a similar Gaussian data setting as in Section~\ref{sec:small}, with the additional difference that the number of nodes is reduced to 20 and the number of seeds is increased to 20.
This study took about 40 minutes to finish on a MacBook Pro 2016 with 3.1 GHz Dual-Core Intel Core i5.
The full \proglang{JSON} specification for this study is found in \emph{paper\_er\_sem\_small.json}.

\begin{figure}[!ht]

\begin{subfigure}[b]{.45\textwidth}
  \centering
  \includegraphics[width=1.0\linewidth]{{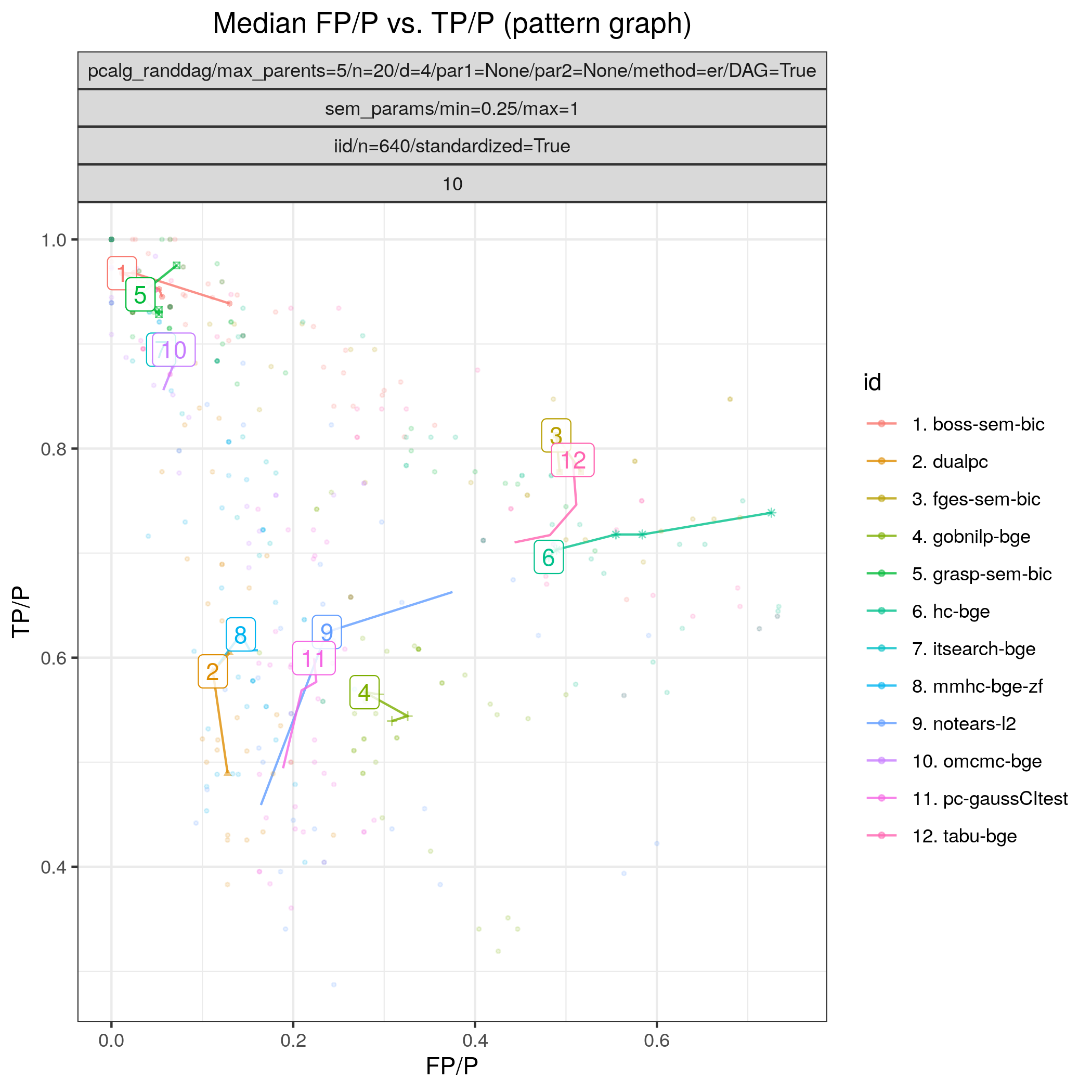}}  
  \caption{}
    \label{fig:repr_roc}
\end{subfigure}
\begin{subfigure}[b]{.45\textwidth}
  \centering
  \includegraphics[width=1.0\linewidth]{{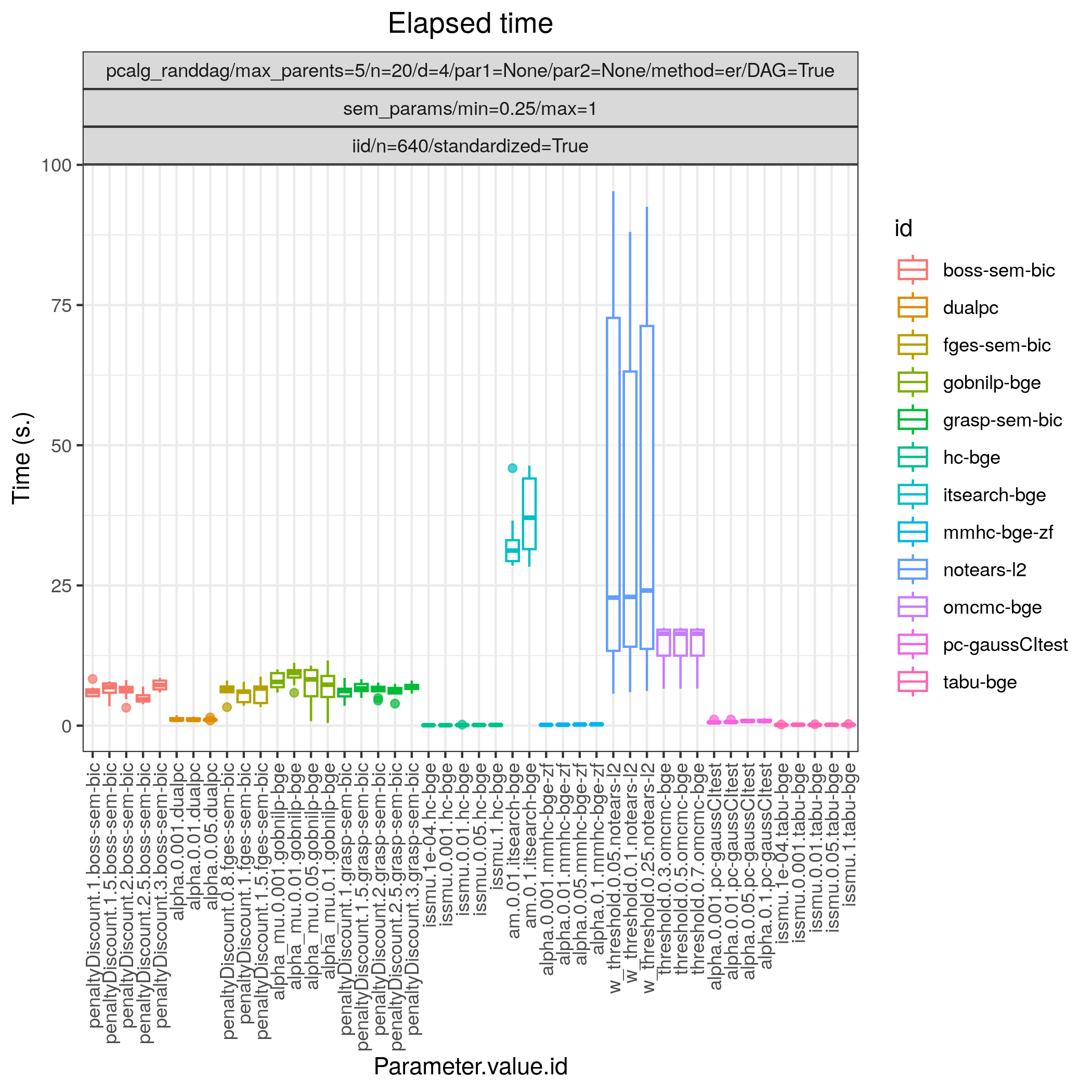}}  
  \caption{}
    \label{fig:repr_time}
\end{subfigure}

\caption{ROC type curves (\ref{fig:repr_roc}) and timing (\ref{fig:repr_time})  from the \texttt{benchmarks} module (see caption of Figure~\ref{fig:small}) evaluated on 20 datasets from a small scale Linear Gaussian SEM with random weights and random DAG. }
\label{fig:rocs_small}
\end{figure}

Figure~\ref{fig:repr_roc} shows the $TP/P$ and $FP/P$ based on pattern graphs and Figure~\ref{fig:repr_time} shows the computational times.
We can directly observe that BOSS (\texttt{boss-sem-bic}), GRaSP (\texttt{grasp-sem-bic}), and the order MCMC (\texttt{omcmc-bge}) have very good performance, though they tend to also have longer computational time.
Apart from this, the results of Figure~\ref{fig:repr_roc} may be partitioned into two regions, FGES (\texttt{fges-sem-bic}), HC (\texttt{hc\_sem-bic}), and Tabu (\texttt{tabu-bge}) having higher values for both $TP/P$ and $FP/P$ and the rest having lower values for both $TP/P$ and $FP/P$.

\subsection{Binary valued Bayesian network with random parameters and random DAG} \label{sec:randbin}
In this example, we study a binary-valued Bayesian network, where both the graph \(\graph\) and the parameters \(\param\) are regarded as random variables.
More specifically, we consider 100 models \(\{(\graph_i,\param_i)\}_{i=1}^{100}\), where each \(\graph_i\) is sampled according to the  Erd\H{o}s-Rényi random DAG model using the \randdag\, module (Appendix~\ref{app:graphs}), where the number of nodes is \(\p=80\), the average number of neighbours (parents) per node is 4 (2) and the maximal number of parents per node is 5. 
The parameters \(\param_i\) are sampled using the \randbn\, module (Appendix~\ref{app:params}) and restricting the conditional probabilities within the range \([0.1, 0.9]\). 
From each model, we draw one dataset \(\rma_i,\) of size \(n=320\) using the \iid\, module. 
The full \proglang{JSON} specification for this study is found in \emph{paper\_er\_bin.json}.

\begin{figure}[!ht]

\begin{subfigure}[b]{.45\textwidth}
  \centering
  \includegraphics[width=1.0\linewidth]{{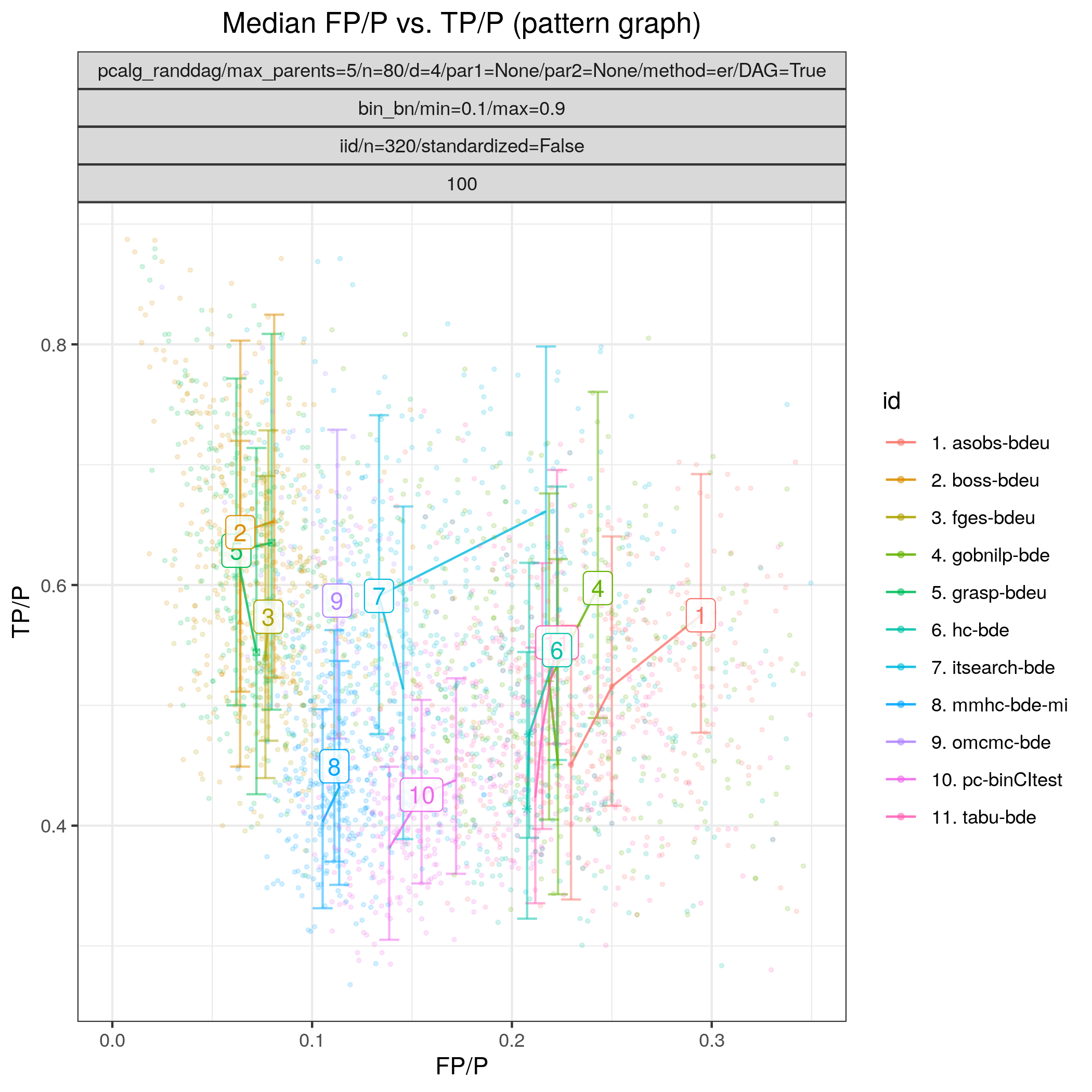}}  
  \caption{Binary valued Bayesian network with random parameters and random DAG.}
    \label{fig:binarybn}
\end{subfigure}
~
\begin{subfigure}[b]{.45\textwidth}
  \centering
  \includegraphics[width=1.0\linewidth]{{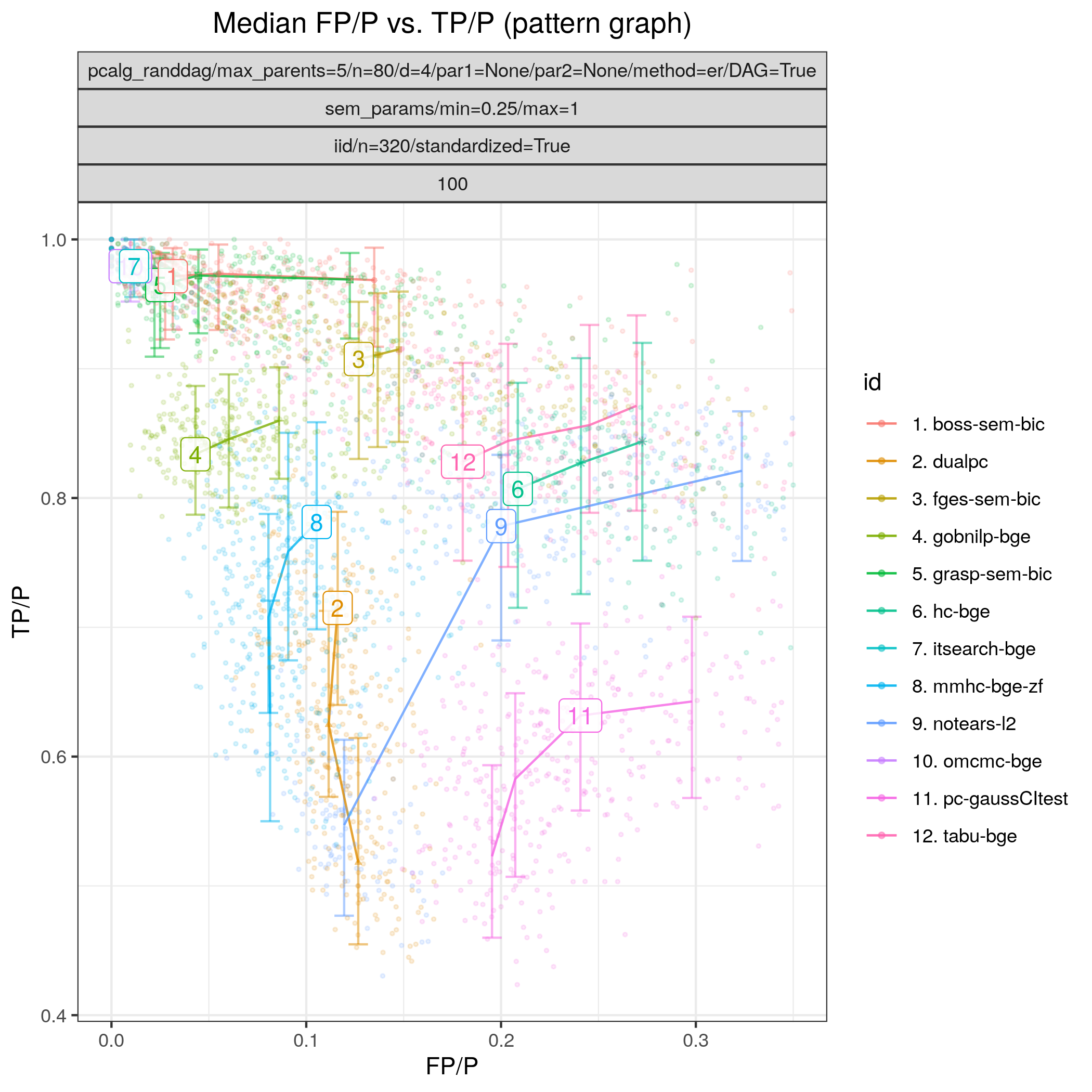}}
\caption{Linear Gaussian SEM with random weights and random DAG.}
  \label{fig:gaussbn}
\end{subfigure}

\begin{subfigure}[b]{.45\textwidth}
  \centering
  \includegraphics[width=1.0\linewidth]{{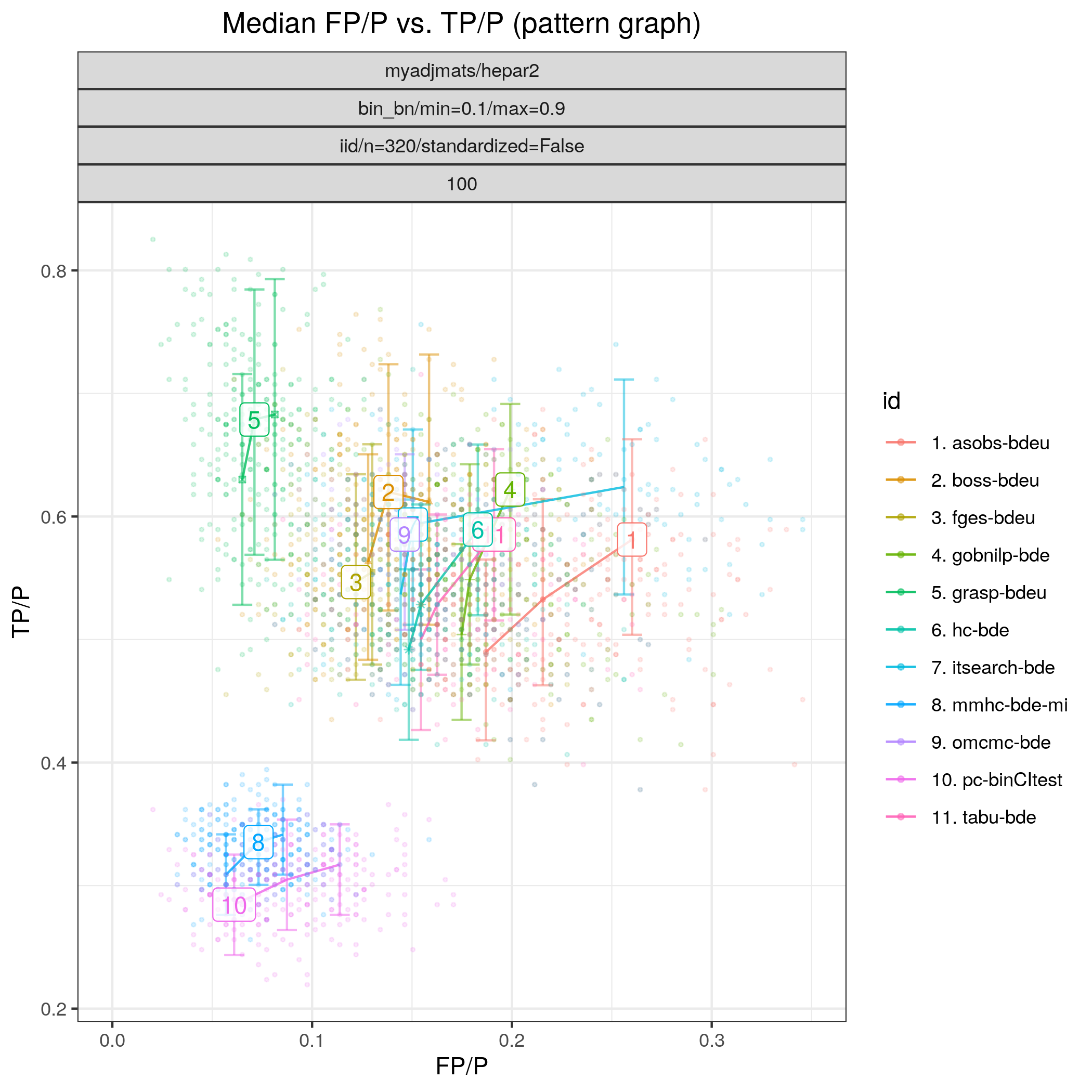}}  
  \caption{Binary valued Bayesian network with random parameters and (fixed) HEPAR II DAG.}
    \label{fig:hepar2bin}
\end{subfigure}
~
\begin{subfigure}[b]{.45\textwidth}
  \centering
  \includegraphics[width=1.0\linewidth]{{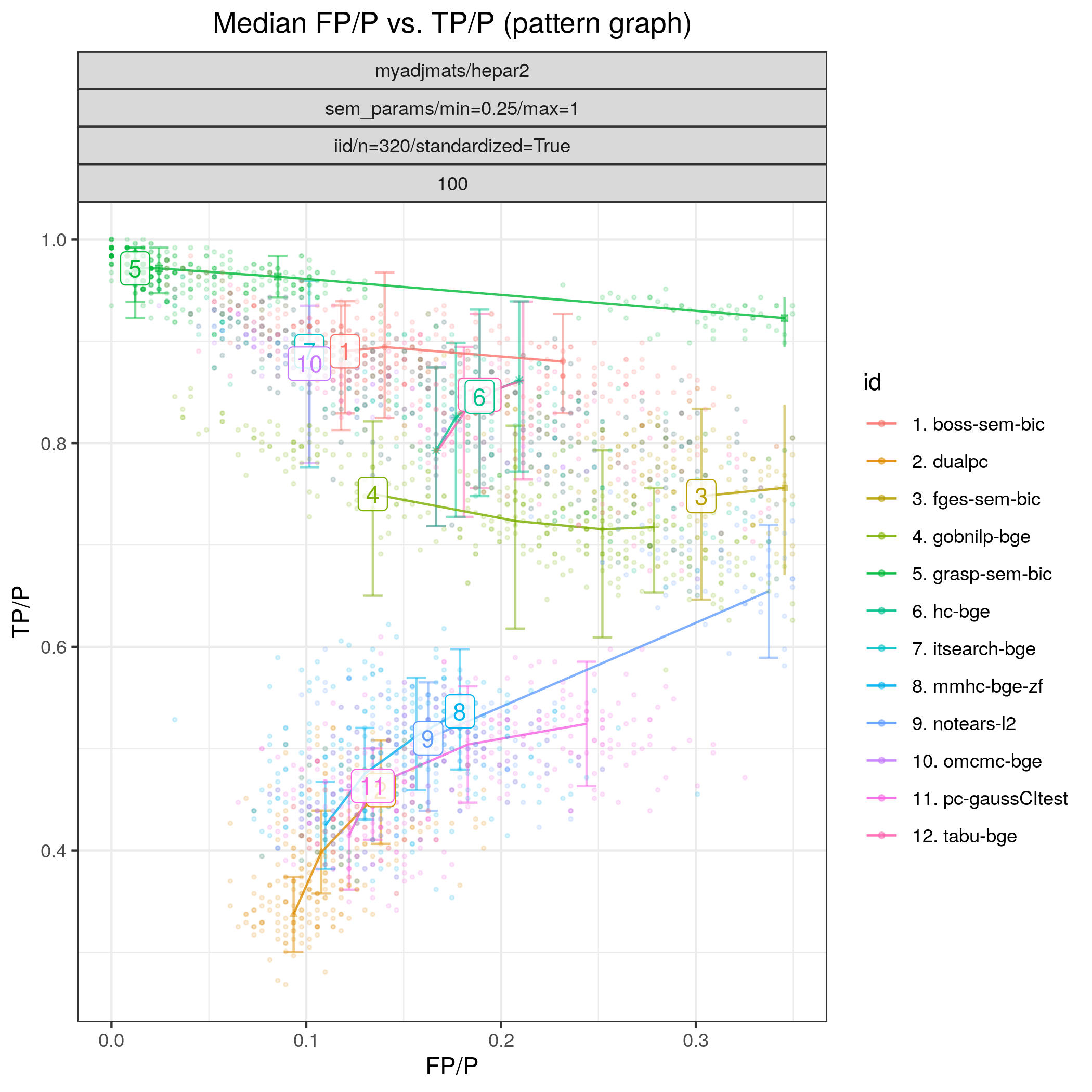}}  
\caption{Linear Gaussian SEM with random weights and (fixed) HEPAR II DAG.}
\label{fig:hepar2gauss}
\end{subfigure}
\caption{ROC type curves (see the caption of Figure~\ref{fig:small} for further details) showing the $FP/P$ and $TP/P$ evaluated on 100 datasets from 4 different Bayesian network models. 
}
\label{fig:rocs}
\end{figure}

Figure~\ref{fig:binarybn} shows the ROC type curves for the algorithms considered for the data generated as described above. 
The algorithms standing out in terms of low SHD in combination with low best median $FP/P$ \((< 0.12)\) and higher best median $TP/P$ \((>0.5)\) are BOSS (\texttt{boss-bdeu}), GRaSP (\texttt{grasp-bdeu}), FGES (\texttt{fges-bdeu}), followed by 
iterative order MCMC (\texttt{omcmc-bde}).

\subsection{Linear Gaussian SEM with random weights and random DAG}
\label{sec:randsem}
In this example, we again study Gaussian random Bayesian networks, of size $p=80$ and with 100 repetitions \(\{(\graph_i,\param_i)\}_{i=1}^{100}\). 
We use the same module configurations as in Sections~\ref{sec:small} and \ref{sec:reproducible} and draw one standardised dataset \(\rma_i\), of size \(n=320\) from each of the models using the \iid\, module.
The full \proglang{JSON} specification for this study is found in \emph{paper\_er\_sem.json}.

Figure~\ref{fig:gaussbn} shows ROC type results for all the algorithms considered for the data generated as described above.
The constraint-based methods PC (\texttt{pc-gaussCItest}) and dual PC (\texttt{dualpc}) have comparable and lower best median $TP/P$ (\(<0.7\)) than the remaining algorithms.
In terms of achieving high $TP/P$ (\(>0.9\)) iterative order MCMC (\texttt{omcmc-bge}) and iterative search MCMC (\texttt{itsearch-bge}) followed by BOSS (\texttt{boss-sem-bic}) and GRaSP (\texttt{grasp-sem-bic})  stand out with near perfect performance, \emph{i.e.}, SHD \(\approx 0\).
Among the other algorithms GOBNILP (\texttt{gobnilp-bge}) performs next best  with $TP/P$ \(\approx 0.85\) and $FP/P$ \(\approx 0.08\) followed by FGES (\texttt{fges-sem-bic}).

\subsection{Bayesian network with random parameters and fixed DAG}

In this example, we study a random binary Bayesian network where the graph \(\graph\) is fixed and the associated parameters \(\param\) are regarded as random.
More specifically, we consider 100 models \(\{(\graph,\param_i)\}_{i=1}^{100}\), where
the graph structure \(\graph\) is that of the well known Bayesian network \emph{HEPAR II} (\emph{hepar2.csv}), first introduced in \cite{onisko2003probabilistic}.
This graph has 70 nodes and 123 edges and has become one of the standard benchmarks for evaluating the performance of structure learning algorithms.
The maximum number of parents per node is 6 and we sample the parameters \(\param_i\) using the \randbn\, module (Appendix~\ref{app:params}), in the same manner as described in Section~\ref{sec:randbin}.
From each model we draw, as before, one dataset \(\rma_i\) of sample size \(n=320\), using the \iid\, module.
The full \proglang{JSON} specification for this study is found in \emph{paper\_hepar2\_bin.json}.

Figure~\ref{fig:hepar2bin} shows the ROC type curves for this scenario. 
The algorithms appear to divide between two groups with respect to their performance in terms of $TP/P$. Constraint-based methods including PC (\texttt{pc-binCItest}), 
and MMHC (\texttt{mmhc-bde-mi}) 
appear to cluster in the lower scoring region ($TP/P$ \(<0.5\)).
Score-based methods on the other hand seem to concentrate in the higher-scoring region ($TP/P$ \(>0.5\)). 
The best performing algorithm is clearly  GRaSP (\texttt{grasp-bdeu}) followed by a group of algorithms including FGES (\texttt{fges-bdeu}), BOSS (\texttt{boss-bdeu}), iterative order MCMC (\texttt{omcmc-bde}), and iterative search MCMC (\texttt{itsearch-bde}), 
all with $FP/P$ \(\approx0.15\). 




\subsection{Linear Gaussian SEM with random weights and fixed DAG} \label{sec:fixeddagsem}
In this example we draw again 100 models \(\{(\graph,\param_i)\}_{i=1}^{100}\), 
where \(\graph\) corresponds to the HEPAR II network (\vals{hepar2.csv}), and \(\param_i\) are the parameters of a linear Gaussian SEM sampled using the \texttt{sem\_params}\, module (Appendix~\ref{app:params}) with the same settings as in Section~\ref{sec:randsem}.
From each of the models \((\graph,\param_i)\), we draw one standardised dataset \(\rma_i\) of size \(n=320\)  (\iid).
The full \proglang{JSON} specification for this study is found in \emph{paper\_hepar2\_sem.json}.

The ROC-type curves of Figure \ref{fig:hepar2gauss} highlight that GRaSP (\texttt{grasp-sem-bic}) again separates itself from the rest of the best algorithms with the more favourable performance. Next follows order MCMC (\texttt{omcmc-bge}), iterative search MCMC (\texttt{itsearch-bge}), and BOSS (\texttt{boss-sem-bic}). 
In terms of best median $TP/P$ (\(>0.8\)) only HC (\texttt{hc-bge}) and Tabu (\texttt{tabu-bge}) display similar performances, while performing considerably worse with respect to median $FP/P$ (\(\approx0.1\) vs \(\approx0.17\)).
\section{Installation and usage}\label{sec:installation}
\subsection{Requirements}
\ttl can run either natively on Linux systems, or through \dkr, which enables it to run also on macOS and Windows.
In either alternative, the first step is to clone the \pkg{git} repository from \url{https://github.com/felixleopoldo/benchpress} and set it to the working directory:
\begin{verbatim}
    $ git clone https://github.com/felixleopoldo/benchpress.git
    $ cd benchpress
\end{verbatim}
\subsubsection{Native (Linux)}\label{sec:usage}
Running the \ttl workflow natively requires \smk \((\geq 7.30.1)\) and \sng \citep{kurtzer2017singularity} to be installed on a Linux system. 
See the documentation of \smk and \sng for specific installation instructions.
If \smk is installed with \pkg{conda} \citep[as suggested in \smk's documentation]{anaconda} in an environment named \emph{e.g.}\ \emph{snakemake},  it should be activated by:
\begin{verbatim}
    $ conda activate snakemake
\end{verbatim}
\subsubsection{Docker (Linux/macOS/Windows)}\label{sec:usage}
For this alternative, the only requirement is \dkr, which can be installed following the instructions on their official website.
The \smk commands are then executed from an interactive shell of the \dkr image \emph{bpimages/snakemake:v7.32.3}, where both \smk and \sng are installed.
The working directory (the \emph{benchpress} directory) is a shared volume and is mounted in the  \emph{/mnt} folder. 
The command is:
\begin{verbatim}
$ docker run -it -w /mnt --privileged -v $(pwd):/mnt bpimages/snakemake:v7.32.3
\end{verbatim}
Windows users should substitute \texttt{\$(pwd)} for the absolute path to the \emph{benchpresss} folder. 
\subsection{Usage}\label{sec:usage}
Configuration files are executed through standard \smk commands, with the \texttt{use-singularity} flag on.
For example, the following command will use all the available cores to run the config file \emph{config/config.json}
\begin{verbatim}
    $ snakemake --cores all --use-singularity --configfile config/config.json
\end{verbatim}
For alternative use cases, \emph{e.g.}\ execution in the cloud or on a grid, we refer to the documentation of \smk.
\newpage
\section{New modules}\label{sec:newalg}
One of the main strengths of \ttl is that the modular design provides a flexible framework to integrate new modules into any of the module sections.
Next, we demonstrate how to add a new module, update the documentation, and push it upstream.

\subsection{Adding new modules}

In this section, we show an example of how to add a new structure learning module.
Adding parameterization and data modules is done similarly.
To get started, copy the template module \texttt{new\_alg} to a new module that we name \texttt{mymod}.
To do this (macOS/Linux) type the following command from the \emph{benchpress} folder:
\begin{verbatim}
    $ cp -r resources/module_templates/new_alg \
    workflow/rules/structure_learning_algorithms/mymod
\end{verbatim}
This will create the new module \texttt{mymod} (in \emph{workflow/rules/structure\_learning\_algorithms/mymod}) which is ready to be used alongside the existing ones.

The module contains files that are needed for the module's functionality and documentation.
It is pre-configured in \emph{rule.smk} (Listing~\ref{code:myalg_rule}, Line~13) to run the template \emph{script.R} (Listing~\ref{code:ralg}), which implements an algorithm that merely samples a random binary symmetric matrix.
The \texttt{snakemake} variable, used in \emph{script.R}, is an object that provides access to the \texttt{input} (Listing~\ref{code:myalg_rule}, Line~4) and \texttt{output} (Listing~\ref{code:myalg_rule}, Line~6) fields of \emph{rule.smk} (Listing~\ref{code:ralg}, Lines~18, 20, and 21) and to the module's \proglang{JSON} object keys through the \texttt{wildcards} list (Listing~\ref{code:myalg_rule}, Lines~7, 10).
Note that, the keys of the \texttt{wildcards} list are directly inherited from the keys of the \proglang{JSON} object used in the config file, see Listing~\ref{json:myalg_json} for an example.
The \texttt{wildcards} list also contains the random seed number (Listing~\ref{code:ralg}, Line 7) stemming from the \texttt{seed\_range} specified in the \texttt{data\_setup} section of the config file.

In the simplest case when adapting the module to your algorithm, you essentially only have to alter the rows between Lines 9-14 in Listing~\ref{code:ralg} and the keys of the \proglang{JSON} object.
The module will by default run in a container based on the \dkr image \emph{\say{bpimages/sandbox:1.0}}, where both \proglang{R} and  \proglang{Python} are installed. 
This may be changed to any suitable image that \smk supports, \emph{e.g.}\ a \dkr image at \emph{Docker Hub}.
To force local execution, which might be desirable when developing a new algorithm, the \texttt{container} field of \emph{rule.smk} (Listing~\ref{code:myalg_rule}, Line~11) should be set to \vals{None}.

Note that the actual algorithm is embedded in the function called  \emph{myalg},  which is passed to the function \code{add\_timeout}. 
This enables a timeout functionality, which writes an empty file if the algorithm has not finished before \code{timeout} seconds, specified in the config file. However, if the algorithm can produce a graph estimate after a pre-specified amount of time, that graph should preferably be written. 


The module template also contains a \proglang{Python} script, \emph{script.py}, which could directly replace \emph{script.R} in \emph{rule.smk} (Listing~\ref{code:myalg_rule}, Line~13).
For other languages and further details about customising \smk rules, we refer to \smk's official documentation.

\begin{listing}
    \begin{minted}[frame=single,
                   framesep=3mm,
                   linenos=true,
                   xleftmargin=21pt,
                   tabsize=4]{Python}
rule:
    name:
        module_name
    input:
        data = alg_input_data()        
    output:
        adjmat = alg_output_adjmat_path(module_name),
        time = alg_output_time_path(module_name),
        ntests = alg_output_ntests_path(module_name)
    container: 
        "docker://bpimages/sandbox:1.0"        
    script:
        "script.R"
 \end{minted}
\caption{\smk rule file \emph{rule.smk} of the \texttt{new\_alg} template.} 
\label{code:myalg_rule}
\end{listing}

\begin{listing}[!th]
    \begin{minted}[frame=single,
                   framesep=3mm,
                   linenos=true,
                   xleftmargin=21pt,
                   tabsize=4]{r}
source("workflow/scripts/utils/helpers.R")

myalg <- function() {    
    data <- read.csv(snakemake@input[["data"]], check.names = FALSE)
    start <- proc.time()[1]
    p <- ncol(data)
    set.seed(as.integer(snakemake@wildcards[["seed"]]))
    
    # Naive algorithm start
    cutoff <- as.numeric(snakemake@wildcards[["cutoff"]])
    adjmat <- matrix(runif(p * p), nrow = p, ncol = p) > cutoff
    adjmat <- 1 * (adjmat | t(adjmat))
    diag(adjmat) <- 0
    # Naive algorithm end
    
    totaltime <- proc.time()[1] - start
    colnames(adjmat) <- names(data) # Get the labels from the data
    write.csv(adjmat, file = snakemake@output[["adjmat"]], 
              row.names = FALSE, quote = FALSE)
    write(totaltime, file = snakemake@output[["time"]])    
    write("None", file = snakemake@output[["ntests"]]) 
}

add_timeout(myalg)
\end{minted}
\caption{The  \proglang{R}-script \emph{script.R} for the \texttt{new\_alg} template module.} 
\label{code:ralg}
\end{listing}

\begin{listing}
    \begin{minted}[frame=single,
                   framesep=3mm,
                   linenos=true,
                   xleftmargin=21pt,
                   tabsize=4]{javascript}
{
    "id": "testing_myalg",
    "cutoff": 0.8,
    "timeout": null
}
 \end{minted}
\caption{Example \proglang{JSON} object for the \texttt{new\_alg} template module.} 
\label{json:myalg_json}
\end{listing}

\subsection{Documentation}
The documentation for a module can easily be generated based on information provided by the files in the modules folder.
The files and their purpose are described next.
The file \emph{schema.json} is a \proglang{JSON} schema for restricting the fields of a module's \proglang{JSON} objects, forcing the user to provide valid input parameters.
It also allows the developer to provide a description of each of the input parameters.
The file \emph{info.json} should be filled with meta-information about the module, in terms of \emph{e.g.}\ version number, links to the documentation, and the type of graph the module supports.
\emph{bibtex.bib} is a \proglang{BibTeX} file where the references, relevant to the module can be added.
\emph{docs.rst} is a documentation file in \proglang{reStructuredText} format that should contain an overview of the module and relevant information.

To generate an \proglang{HTML} version of the documentation using \emph{sphinx}, the following commands should be executed from the \emph{benchpress} directory (macOS/Linux):
\begin{Verbatim}[samepage=true]
    $ pip install -r docs/source/requirements.txt
    $ cd docs
    $ chmod +x render_docs.sh
    $ ./render_docs.sh
    $ make html
\end{Verbatim}
The documentation can be seen by opening \emph{docs/build/html/index.html} with a web browser.

\subsection{Contributing}
The instructions above show how to integrate a new module into \ttl.
To add a  module to the \ttl official repository, the module should run on a \dkr or \sng image available at \emph{e.g.} Docker Hub.
Publishing modules (or any contributions) to \ttl is done by creating a so-called fork of the \ttl's \emph{GitHub} repository (\url{https://github.com/felixleopoldo/benchpress}) and creating a so-called pull request.  
    
\subsection{Reproducibility}

All the pieces of information used to build the benchmarks in \ttl are saved as files that are easily accessible.
For all the simulations, \ttl uses fixed random seeds to ensure that the code will produce the same results again.
Using \dkr and \sng makes it easy to reproduce results on essentially any modern computer.
When a new module is published, we encourage the developer to attach one or more config files that highlight important properties of the modules along with a few example output figures in \emph{docs.rst}.



\section{Conclusions}
\ttl provides a novel \smk workflow for scalable and reproducible execution and benchmarking of structure learning algorithms.
\smk's support for running containerized software through \sng together with the simple data and graph representation enables \ttl to compare algorithms implemented in different programming languages and run them without requiring unnecessary system privileges or installation of individual module dependencies.
\ttl is the first software of its kind for structure learning in many aspects, including scalability and reproducibility, and perhaps most importantly from the fact that it can directly incorporate existing software.
This can potentially save researchers a large amount of unnecessary work and provide benchmarks that would never be implemented otherwise. 
Likewise, it could facilitate necessary benchmarks for new algorithms, and the sharing of the results in a standardised format.
In its current version \ttl already implements \nalgs of the state-of-the-art learning algorithms, as well as several modules for generating data models and evaluating performance.
As \ttl is built in a completely modular form \smk allows for seamless scaling of computations over multiple cores, grids or servers without any extra additional effort. 
In addition, it is straightforward to integrate new modules into the workflow.
Even though the \ttl project focusses so far on structure learning for graphical models,  
we also see the potential in extending \ttl to evaluate more general estimation procedures, also for other statistical models.


\section*{Acknowledgments}
We would like to thank all the developers and researchers who made their software available open source.
A special thanks to James Cussens and Mohamad Elmasri for their valuable feedback and their contributions towards expanding the package.



\appendix
\section{Graph metrics and data formats}
\label{appendix}
\subsection{Metrics}\label{appendix:metrics}
This section describes metrics for comparing graphs.
We let \(\graph=(\graphnodeset,\graphedgeset)\) and \(\graph'=(\graphnodeset',\graphedgeset')\) denote the true and the estimated graph, respectively. 

\subsubsection{Structural Hamming distance} 
The \emph{structural Hamming distance} (SHD) is one of the most commonly used metrics to compare graphs.
It describes the number of changes, in terms of adding, removing or reversing edges or their directions, needed to transform \(\graphedgeset\) into \(\graphedgeset'\).

\subsubsection{True and false positive rates for mixed graphs}
The following metric quantifies the difference between two \emph{mixed graphs}, which may have a combination of directed and undirected edges.
We let $TP$ and $FP$ be the true and also positive edge rates, but for directed edges,
we include errors in their direction (wrong direction or directed where it should be undirected, or vice versa) as half a false positive ($FP$) and half a false negative ($FN$).

We assign to every edge \(e \in \graphedgeset'\) the true positive score \(TP(e)=1\) if \(e\) is contained in \(\graphedgeset\) with the same orientation (\(e\) being undirected in both \(\graphedgeset\) and \(\graphedgeset'\) or \(e\) having the same direction in both \(\graphedgeset\) and \(\graphedgeset'\)), \(TP(e)=1/2\) if \(e\) is contained in \(\graphedgeset\) with a different orientation (or undirected when should be directed, or vice versa), otherwise \(TP(e)=0\).

The false positive score \(FP(e)\) is defined analogously
for every edge \(e \in \graphedgeset'\). 
\(FP(e)=1\) if no orientation of \(e\)  is contained in \( \graphedgeset\)  
and \(FP(e)=1/2\) if \(e\) is contained in \( \graphedgeset\) with a different orientation (or undirected when should be directed, or vice versa), otherwise \(FP(e)=0\).

The total true and false positive edges ($TP$ and $FP$) are obtained as the sum of the individual edge scores, \emph{i.e.}
\begin{align*}
    TP := \sum_{e\in \graphedgeset '} TP(e), \quad FP := \sum_{e\in \graphedgeset'} FP(e)
\end{align*}
and the false negative edges ($FN$) are defined as
\begin{align*}
    FN:= P - TP,
\end{align*}
where \(P:=|\graphedgeset|\) denotes the total number of edges in \(\graph\).
Note that $TP$, $FP$, and $FN$ reduce to the ordinary true and false positives and false negatives, respectively,  when all edges are undirected in both \(\graph\) and \(\graph'\).

We often consider the scaled true and false positive rates $FP/P$ and $TP/P$, since
the SHD, also for mixed graphs, can easily be determined in a ROC-type figure as the scaled \emph{Manhattan distance} between the points (0,1) and $(TP/P, FP/P)$, \emph{i.e.}
\begin{align*}
    SHD/P= 1-TP/P + FP/P.
\end{align*}
\subsubsection{$F$ score}
The \emph{precision} ($PR$) and \emph{recall} ($RE$) metrics are defined as
\begin{align*}
    PR := \frac{TP}{TP+FP},\quad RE:=\frac{TP}{TP+FN}.
\end{align*}
For any \(\beta\ge0\) the \(F_\beta\) score is defined as
\begin{align*}
        F_\beta := (1+\beta^2)\frac{PR \cdot RE}{\beta^2PR + RE}.
\end{align*}
\ttl uses \(\beta=1\) as default which simplifies to
\begin{align*}
        F_1 = \frac{PR \cdot RE}{PR + RE}= \frac{2TP}{2TP +FP +FN}.
\end{align*}

\subsection{Data formats} \label{app:formats}
Throughout this section for simplicity we consider a four dimensional graphical model where the nodes are labeled as \emph{a,b,c} and \emph{d}.

\subsubsection{Data set}
Observations should be stored as row vectors in a matrix, where the columns are separated by commas. 
The first row should contain the labels of the variables and if the data is categorical, the second row should contain the cardinality (number of levels) of each variable.

Below is a formatting example of two samples of a categorical distribution where the cardinalities are 2,3,2, and 2.
\begin{verbatim}
    a,b,c,d
    2,3,2,2
    1,2,0,1
    0,1,1,1
\end{verbatim}
An example showing of two samples from continuous distribution is shown below.
\begin{verbatim}
    a,b,c,d
    0.2,2.3,5.3,0.5
    3.2,1.5,2.5,1.2
\end{verbatim}
\subsubsection{Adjacency matrix}
A graph \(\graph=(E,V)\) is represented by  its adjacency matrix \(M\), where \(M_{ij}=1\) if \((i,j)\in \graphedgeset\) and  \(M_{ij}=0\) if \((i,j)\notin \graphedgeset\).
An undirected graph is represented by a symmetric matrix.

Below is an example undirected graph \(\graph=(\graphnodeset, \graphedgeset)\), where \(\graphedgeset = \{(a,b), (a,c), (c,d)\}\) are interpreted as un-ordered pairs (un-directed edges).
\begin{verbatim}
    a,b,c,d
    0,1,1,0
    1,0,0,0
    1,0,0,1
    0,0,1,0
\end{verbatim}
If \(\graph\) is directed
the adjacency matrix is asymmetric as below.
\begin{verbatim}
    a,b,c,d
    0,1,1,0
    0,0,0,0
    0,0,0,1
    0,0,0,0
\end{verbatim}

\subsubsection{MCMC trajectory}
 When the output of the algorithm is a Markov chain of graphs, we store the output in a compact form by tracking only the changes when moves are accepted, along with the corresponding time index and the score of the resulting graph after acceptance (not the score difference). 

Additionally, in the first two rows the labels of the variables, which should be read from the data matrix, are recorded. 
Specifically, the first row (index -2) contains edges from the first variable to each of the rest in the \texttt{added} column, where a dash (\texttt{-}) symbolises an undirected edge, and a right arrow (\texttt{->}) a directed edge. 
The \texttt{score} column is set to 0 and \texttt{removed} is set to \texttt{[]}. 
 The second row (index -1) has the same edges in the \texttt{removed} column, while the \texttt{score} column is set to 0 and \texttt{added} is set to \texttt{[]}. 
 The third row (index 0) contains all the vertices in the starting graph along with its score in the \texttt{score} column and \texttt{[]} in the \texttt{removed} column.
 
 Below is an example of a trajectory of undirected graphs \(\graph_0,\graph_1,\dots ,\graph_{89}\), 
 where \(E_i=\{(b,c), (a,d)\}\) for \(i=0,\dots,33\), \(E_i=\{(a,d)\}\) for \(i=34,\dots,88\) and \(E_i=\{(c,d), (a,d)\}\) for \(i=89\).
 \begin{verbatim}
    index,score,added,removed
    -2,0.0,[a-b;a-c;a-d],[]
    -1,0.0,[],[a-b;a-c;a-d]
    0,-2325.52,[b-c;a-d],[]
    34,-2311.94,[],[b-c]
    89,-2310.81,[c-d],[]
\end{verbatim} 

\section{Benchpress modules}  \label{app:modules}

\subsection{Graph modules}\label{app:graphs}
\noindent \emph{Random graph} (\randdag) \vspace{0.2cm}\\
The \randdag\, module samples random undirected graphs and DAGs using the \code{randDAG} function from the \pkg{pcalg} package \citep{kalisch2012causal}, with the extra option of restricting the maximal number of parents per node. 
\\\\
\noindent \emph{Fixed graph} (\texttt{fixed\_graph}) \vspace{0.2cm}\\
A fixed graph is represented as an adjacency matrix and should be formatted as specified in Appendix~\ref{appendix}. 
The file should be stored with the \emph{.csv} extension in the directory \emph{resources/adjmat/myadjmats/}.

\subsection{Parameters modules}\label{app:params}
\noindent \emph{Random binary Bayesian network} (\randbn) \vspace{0.2cm}\\
%
This module samples the conditional probability tables of a binary Bayesian network (only binary variables). 
For each variable \(\rva_i\) and parent configuration \(\pa(\va_i)\) 
\begin{align*}
    \probf(\rva_i=0 | \pa(\va_i) ) \sim \mathrm{Unif}([a, b]),
\end{align*}
where  \((a,b) \in [0,1]^2, a<b\) and \( \mathrm{Unif}(c)\) denotes the uniform distribution on the range \(c\).
\\\\
\noindent \emph{Random linear Gaussian Bayesian network} (\texttt{sem\_params}) \vspace{0.2cm}\\
This module samples the weight matrix \(W\) of a Gaussian linear structural equation model (SEM) of the form
\begin{align}
\rva_i=\sum_{j:\rva_j\in \pa(\rva_i)} W_{ij}\rva_j + \rvc_i,
\label{eq:sem}
\end{align}
where \(\rvc_i\sim \mathcal N(\mu, \sigma^2)\) and elements of \(W\) are distributed as
\begin{align}\label{eq:sem_weights}
W_{ij} \sim 
\begin{cases}
\mathrm{Unif}([a, b])\mathrm{Unif}(\{-1,1\}) & \text{ if }(i, j) \in \graphedgeset\\
0 & \text{ otherwise.}
\end{cases}
\end{align}
%
\\\\
\noindent \emph{Fixed parameters} (\texttt{fixed\_params}) \vspace{0.2cm}\\
Two types of fixed parameters are currently supported and described below.

\begin{itemize}
\item \emph{bn.fit object}: A Bayesian network object in the \proglang{R}-package \pkg{bnlearn} \citep{JSSv035i03} is an instance of the \emph{bn.fit} class and contains both the DAG and corresponding parameters.
Motivated by the modular design of \ttl, a \emph{bn.fit} object may be used to specify only the parameters  of a Bayesian network, by storing it in \emph{.rds} format in \emph{resources/parameters/myparams/bn.fit\_networks/}.
The adjacency matrix of the DAG should be stored in the folder for \texttt{fixed\_graphs} and \texttt{graph\_id} should be set to the filename.

\item \emph{SEM parameters matrix}: A matrix defining the weights \(W\) in a SEM as defined in \eqref{eq:sem} can be stored  in \nolinebreak \emph{resources/parameters/myparams/sem\_params/} with the \emph{.csv} extension, formatted as the adjacency matrix specified in Appendix~\ref{appendix}.
\end{itemize}

\noindent \emph{Gaussian graph intra-class model} (\texttt{trilearn\_intraclass}) \vspace{0.2cm}\\
This module specifies the covariance matrix $\covmat$ 
of a zero-mean Gaussian graphical model by the solution of the following matrix completion problem
\begin{align*}
        \covmat_{ij} &= \begin{cases}
        \sigma^2, &\text{ if } i=j\\
        \rho\sigma^2, &\text{ if } (i,j) \in \graphedgeset
    \end{cases}\\
    (\covmat^{-1})_{ij} &=0 \quad \quad\quad \text{ if } (i,j) \notin \graphedgeset,
\end{align*}
where \(\sigma^2>0\) and \(\rho \in [0,1]\) denote the variance and correlation coefficient, respectively.
\\\\
\noindent \emph{Hyper Dirichlet} (\texttt{trilearn\_hyper-dir}) \vspace{0.2cm}\\
This module samples the parameters of a categorical decomposable model from the \emph{hyper Dirichlet} distribution \citep{dawid1993}, with a specified concentration parameter and number of levels per variable.
\\\\
\noindent \emph{Inverse G-Wishart} (\texttt{bdgraph\_rgwish}) \vspace{0.2cm}\\
This modules samples the precision matrix of a Gaussian graphical model from the \emph{G-Wishart} distribution \citep{dawid1993,Atay-Kayis01062005, lenkoski2013direct} using the \code{rgwish} function from the \pkg{BDgraph} package \citep{JSSv089i03}.
The clique-wise scale matrices are fixed to the identity matrix while the degrees of freedom and a threshold value for the convergence of the sampling algorithm are specified by the user.

\subsection{Data modules}\label{app:data}
\noindent \emph{Fixed data sets} (\texttt{fixed\_data}) \vspace{0.2cm}\\
The are two ways of providing fixed data sets.
The first option is to place data files directly in the directory \emph{resources/data/mydatasets/} with the \emph{.csv} extension, formatted according to Appendix~\ref{appendix}.
The second option is to place the files in a sub directory of \emph{resources/data/mydatasets/}. 
In the latter case, the \texttt{data\_id} field in the \texttt{benchmarks\_setup} section should be the name of the directory and all the files in it will be considered for evaluation.
\\\\
\noindent \emph{Independent identically distributed (i.i.d.) samples} (\iid) \vspace{0.2cm}\\
An object of the \iid\, module will draw a specified (\texttt{n}) number of independent samples from a specified model. 
Continuous data may be standardized by setting \texttt{standardized} to \vals{true}.
See \cite{reisach2021beware} for a discussion about standardising data in a structure learning context.

\subsection{Structure learning algorithms}
This section contains a short summary of the structure learning modules that are used in the simulation study of Section~\ref{sec:case_study}.

\begin{table}[!ht]
    \small
    \centering
    \begin{tabular}{llll}
        \textbf{Algorithm} & \textbf{Graph} & \textbf{Package} & \textbf{Module} \\ \hline
        ANM & DAG & \pkg{gCastle} & \texttt{gcastle\_anm} \\ 
        ASOBS & DAG & \pkg{r.blip} & \texttt{rblip\_asobs} \\ 
        BDgraph & UG & \pkg{BDgraph} & \texttt{bdgraph} \\ 
        BOSS & CPDAG & \pkg{causal-cmd} & \texttt{tetrad\_boss} \\ 
        Chordal graph samplers & DG & ~ & \texttt{athomas\_jtsamplers} \\ 
        CORL & DAG & \pkg{gCastle} & \texttt{gcastle\_corl} \\ 
        Corrmat thresh & UG & \pkg{Benchpress} & \texttt{corr\_thresh} \\ 
        Direct LINGAM & DAG & \pkg{gCastle} & \texttt{gcastle\_direct\_lingam} \\ 
        Dual PC & CPDAG & \pkg{dualPC} & \texttt{dualpc} \\ 
        FAS & DAG & \pkg{causal-cmd} & \texttt{tetrad\_fas} \\ 
        FASK & DAG & \pkg{causal-cmd} & \texttt{tetrad\_fask} \\ 
        Fast IAMB & DAG & \pkg{bnlearn} & \texttt{bnlearn\_fastiamb} \\ 
        FGES & CPDAG & \pkg{causal-cmd} & \texttt{tetrad\_fges} \\ 
        FOFC & DAG & \pkg{causal-cmd} & \texttt{tetrad\_fofc} \\ 
        FTFC & DAG & \pkg{causal-cmd} & \texttt{tetrad\_ftfc} \\ 
        GAE & DAG & \pkg{gCastle} & \texttt{gcastle\_gae} \\ 
        GIES & CPDAG & \pkg{pcalg} & \texttt{pcalg\_gies} \\ 
        GOBNILP & DAG & \pkg{GOBNILP} & \texttt{gobnilp} \\ 
        GOLEM & DAG & \pkg{gCastle} & \texttt{gcastle\_golem} \\ 
        GraNDAG & DAG & \pkg{gCastle} & \texttt{gcastle\_grandag} \\ 
        Graphical Lasso & UG & \pkg{scikit-learn} & \texttt{sklearn\_glasso} \\ 
        GRaSP & CPDAG & \pkg{causal-learn} & \texttt{causallearn\_grasp} \\ 
        GRaSP & CPDAG & \pkg{causal-cmd} & \texttt{tetrad\_grasp} \\ 
        Grow-shrink & DAG & \pkg{bnlearn} & \texttt{bnlearn\_gs} \\ 
        GrUES & UG & \pkg{gues} & \texttt{grues} \\ 
        GSP & DAG & \pkg{CausalDAG} & \texttt{causaldag\_gsp} \\ 
        H2PC & DAG & \pkg{bnlearn} & \texttt{bnlearn\_h2pc} \\ 
        HC & DAG & \pkg{bnlearn} & \texttt{bnlearn\_hc} \\ 
        HPC & DAG & \pkg{bnlearn} & \texttt{bnlearn\_hpc} \\ 
        IAMB & DAG & \pkg{bnlearn} & \texttt{bnlearn\_iamb} \\ 
        IAMB-FDR & DAG & \pkg{bnlearn} & \texttt{bnlearn\_iambfdr} \\ 
        ICALiNGAM & DAG & \pkg{gCastle} & \texttt{gcastle\_ica\_lingam} \\ 
        INTER-IAMB & DAG & \pkg{bnlearn} & \texttt{bnlearn\_interiamb} \\ 
        Iterative search & DAG & \pkg{BiDAG} & \texttt{bidag\_itsearch} \\ 
        LINGAM & DAG & \pkg{causal-cmd} & \texttt{tetrad\_lingam} \\ 
        MCSL & DAG & \pkg{gCastle} & \texttt{gcastle\_mcsl} \\ 
        MMHC & DAG & \pkg{bnlearn} & \texttt{bnlearn\_mmhc} \\ 
        MMPC & DAG & \pkg{bnlearn} & \texttt{bnlearn\_mmpc} \\ 
        NO TEARS & DAG & \pkg{gCastle} & \texttt{gcastle\_notears} \\ 
        NO TEARS low rank & DAG & \pkg{gCastle} & \texttt{gcastle\_notears\_low\_rank} \\ 
        NO TEARS non-linear & DAG & \pkg{gCastle} & \texttt{gcastle\_notears\_nonlinear} \\ 
        Order MCMC & DAG & \pkg{BiDAG} & \texttt{bidag\_order\_mcmc} \\ 
        Parallel DG & DG & \pkg{parallelDG} & \texttt{paralleldg} \\ 
        Particle Gibbs & DG & \pkg{trilearn} & \texttt{trilearn\_pgibbs} \\ 
        Partition MCMC & DAG & \pkg{BiDAG} & \texttt{bidag\_partition\_mcmc} \\ 
        PC & CPDAG & \pkg{bnlearn} & \texttt{bnlearn\_pcstable} \\ 
        PC & DAG & \pkg{gCastle} & \texttt{gcastle\_pc} \\ 
        PC & CPDAG & \pkg{pcalg} & \texttt{pcalg\_pc} \\ 
        PC-ALL & DAG & \pkg{causal-cmd} & \texttt{tetrad\_pc-all} \\ 
        Precmat thresh & UG & \pkg{Benchpress} & \texttt{prec\_thresh} \\ 
        Psi-leaner & UG & \pkg{equSA} & \texttt{equsa\_psilearner} \\ 
        RL & DAG & \pkg{gCastle} & \texttt{gcastle\_rl} \\ 
        RSMAX2 & DAG & \pkg{bnlearn} & \texttt{bnlearn\_rsmax2} \\ 
        S-I HITON-PC & DAG & \pkg{bnlearn} & \texttt{bnlearn\_sihitonpc} \\ 
        Tabu & DAG & \pkg{bnlearn} & \texttt{bnlearn\_tabu} \\ 
    \end{tabular}

      \caption{Currently available structure learning modules.}
        \label{tab:algorithms}
\end{table}

\noindent \emph{Peter and Clark} (\texttt{pcalg\_pc})\vspace{0.2cm}\\
The \emph{Peter and Clark} (PC) algorithm \citep{spirtes1991algorithm}, is a constraint based method consisting of two main steps.
The first step is called the \emph{adjacency search} and amounts to finding the undirected skeleton of the DAG through conditional independence tests.
The second step consists of estimating a CPDAG from the skeleton and the previous tests.
\\\\
\noindent \emph{Dual PC} (\texttt{dualpc})\vspace{0.2cm}\\
 The \emph{dual PC} algorithm \citep{giudice2021dual} is an alternative scheme to carry out the conditional independence tests within the PC algorithm for Gaussian data, by leveraging the inverse relationship between covariance and precision matrices. 
 The algorithm exploits block matrix inversions on the covariance and precision matrices to simultaneously perform tests on partial correlations of complementary (or dual) conditioning sets.
 Simulation studies indicate that the dual PC algorithm outperforms the classic PC algorithm both in terms of run time and in recovering the underlying network structure.
\\\\
\noindent \emph{Hill climbing} (\texttt{bnlearn\_hc})\vspace{0.2cm}\\
\emph{Hill climbing} (HC) is a score-based algorithm which starts with a DAG with no edges and adds, deletes or reverses edges in a greedy manner until an optimum is reached \citep{russell2002artificial,scutari2019learning}.
\\\\
\noindent \emph{Tabu} (\texttt{bnlearn\_tabu})\vspace{0.2cm}\\
\emph{Tabu} is a less greedy version of the HC algorithm allowing for non-optimal moves that might be beneficial from a global perspective to avoid local maxima \citep{russell2002artificial, scutari2019learning}.
\\\\
\noindent \emph{Best Order Score Search} (\texttt{tetrad\_boss})\vspace{0.2cm}\\
\emph{Best Order Score Search} (BOSS) \citep{DBLP:journals/corr/abs-2108-10141,andrews2023fast}  is a permutation-based algorithm stemming from  the \emph{Ordering Search} of \cite{teyssier2012ordering} and the \emph{Sparsest Permutation} algorithm (SP)  of \cite{10.1002/sta4.183} as in the \emph{Greedy Sparsest Permutation} algorithm GSP of \cite{10.1093/biomet/asaa104}. 
BOSS gives results as accurate as SP but for much larger and denser graphs. 
It is more accurate for two reason: (a) It assumes the so-called  \emph{brute faithfuness} assumption, which is weaker than faithfulness, and (b) it uses a different traversal of permutations than the depth-first traversal used by GSP, obtained by taking each variable in turn and
moving it to the position in the permutation that optimizes the model score.
\\\\
\noindent \emph{Fast greedy equivalent search} (\texttt{tetrad\_fges})\vspace{0.2cm}\\
\emph{Fast greedy equivalent search} (FGES) is a score based method based on the  the greedy equivalence search (GES) of \cite{meek1997graphical}.
This algorithm operates on the space of CPDAG's \citep{chickering2002optimal}.
Its complexity is polynomial in the number of nodes.
The FGES is asymptotically correct under the assumption that there are no unmeasured confounders \citep{ogarrio2016hybrid}, a condition required for most algorithms with convergence guarantees.
\\\\
\noindent \emph{Greedy relaxations of the sparsest permutation algorithm} (\texttt{tetrad\_grasp})\vspace{0.2cm}\\
    This is a method that exploits permutation reasoning to search for directed acyclic causal models, like the \emph{Ordering Search} of \cite{teyssier2012ordering} and  GSP of \cite{10.1093/biomet/asaa104}. 
The algorithm extends these methods by a permutation-based operation called \emph{tuck}, and develops a class of algorithms, namely \emph{Greedy relaxations of the sparsest permutation} (GRaSP) \citep{pmlr-v180-lam22a}, that are computationally efficient and pointwise consistent under increasingly weaker assumptions than faithfulness.

\noindent \emph{Order Markov chain Monte Carlo} (\texttt{bidag\_order\_mcmc})\vspace{0.2cm}\\
This technique relies on a Bayesian perspective on structure learning and uses the posterior probability of graphs as a score. 
To overcome the limitation of simple structure-based MCMC schemes, \cite{Friedman2003} turned to a score defined as the sum of the posterior scores of all DAG which are consistent with a given topological ordering of the nodes. One can then run a Metropolis-Hasting algorithm to sample from the distribution induced by the order score, and later draw a DAG consistent with the order.
This strategy substantially improves convergence with respect to earlier structure MCMC scheme, though it unfortunately produces a biased sample on the space of DAGs. The bias can be removed by operating on the space of ordered partitions instead \citep{doi:10.1080/01621459.2015.1133426}. 
The implementation considered in \ttl is a hybrid version with the sampling performed on a restricted search space initialised with constraint-based testing and improved with a score-based search \citep{doi:10.1080/10618600.2021.2020127}. 
\\\\
\noindent \emph{Max-min hill-climbing} (\texttt{bnlearn\_mmhc})\vspace{0.2cm}\\
 \emph{Max-min hill-climbing} (MMHC) is a hybrid method which first estimates the skeleton of a DAG using an algorithm called \emph{Max-Min Parents and Children} and then performs a  greedy hill-climbing search to orient the edges with respect to a Bayesian score \citep{tsamardinos2006max}.
It is a popular approach used as standard benchmark and also well suited for high-dimensional domains.
\\\\
\noindent \emph{Globally optimal Bayesian network learning using integer linear programming} (\texttt{gobnilp})\vspace{0.2cm}\\
A score based method using integer linear programming (ILP) for learning an optimal DAG for a Bayesian network with limit on the maximal number of parents for each node \citep{cussens2012bayesian}.
It is a two-stage approach where candidate parent sets for each node are discovered in the first phase and the optimal sets are determined in a second phase.
\\\\
\noindent \emph{Acyclic selection ordering-based search} (\texttt{rblip\_asobs})\vspace{0.2cm}\\
A score-based two-phase algorithm where the first phase aims to identify the possible parent sets, \cite{scanagatta2015learning, scanagatta2018approximate}.
The second phase performs an optimisation on a modification of the space of node orders introduced in \cite{teyssier2012ordering}, allowing edges from nodes of higher to lower order, provided that no cycles are introduced.
\\\\
\noindent \emph{Iterative search} (\texttt{bidag\_itsearch})\vspace{0.2cm}\\
This is a hybrid score-based optimisation technique based on Markov chain Monte Carlo schemes \citep{suter2021bayesian, doi:10.1080/10618600.2021.2020127}. The algorithm starts from a skeleton obtained through a fast method (\emph{e.g.}\ a constraint based method, or GES). Then it performs score and search on the DAGs belonging to the space defined by the starting skeleton. To correct for edges which may be missed, the search space is iteratively expanded to include one additional parent for each variable from outside the current search space. The score and search phase relies on an MCMC scheme producing a chain of DAGs from their posterior probability given the data.
\\\\
\noindent \emph{No tears} (\texttt{gcastle\_notears})\vspace{0.2cm}\\
This score-based method recasts the combinatorial problem of estimating a DAG into a purely continuous non-convex optimization problem over real matrices with a smooth constraint to ensure acyclicity \citep{zheng2018dags}.




\subsection{Evaluation modules}
\noindent \emph{Standard benchmarking metrics} (\texttt{benchmarks}) \vspace{0.2cm}\\
The relative performance of algorithms may differ depending on the evaluation metric, and no single metric is generally preferred.
Therefore, to get an overall picture of the performance of an algorithm, the \texttt{benchmarks} module supports different metrics.
The \texttt{benchmarks} module provides standard benchmarking metrics in terms of computational time, $F_1$, $FN$, $TP$, and $FP$, \emph{etc.} (see Appendix~\ref{appendix:metrics} for definitions). 
The results are saved systematically in CSV format, which can be analysed using any program. 

In addition to the CSV summaries, \ttl also provides visual summarises in terms of \emph{e.g.}, box-plots and receiver operating characteristics (ROC) type curves using the \proglang{R}-package \pkg{ggplot2} \citep{tidyverse} (see \emph{e.g.}\ Figure~\ref{fig:small_roc} and Figure~\ref{fig:rocs}). 
Since the true graphs are needed for evaluations, this module works for data scenarios (II-V).
\\\\
\noindent \emph{Pair-wise plots of the data} (\texttt{ggally\_ggpairs}) \vspace{0.2cm}\\
The \texttt{ggally\_ggpairs} module produces pair-wise plots of the data using the \code{ggpairs} function \citep{emerson2013generalized} from the \proglang{R}-package \pkg{GGally}  (see Figure~\ref{fig:sachs_pairs} for an example).
\\\\
\noindent \emph{Plot estimated graphs} (\texttt{graph\_plots}) \vspace{0.2cm}\\
The \texttt{graph\_plots} module plots and saves the estimated graphs and adjacency matrices (see Figure~\ref{fig:sachs_adjmat} and \ref{fig:sachs_graphest} for examples). 
If the true graph is available it also compares the true to the estimated graphs using \code{graphviz.compare} function from the \pkg{bnlearn} package (see Figure~\ref{fig:sachs_compare} for an example).
\\\\
\noindent \emph{Plot true graphs} (\texttt{graph\_true\_plots}) \vspace{0.2cm}\\
The \texttt{graph\_true\_plots} module plots the true underlying graphs and corresponding adjacency matrices. 
\\\\
\noindent \emph{Statistics for true graphs} (\texttt{graph\_true\_stats}) \vspace{0.2cm}\\
The \texttt{graph\_true\_stats} module computes properties of the true underlying graphs and stored in a CSV file and plotted.
See Figure~\ref{fig:small_density} for an example plot from this module.
\\\\
\noindent \emph{MCMC mean graphs} (\texttt{mcmc\_heatmaps}) \vspace{0.2cm}\\
For Bayesian inference it is customary to use MCMC methods to simulate a Markov chain of graphs  \(\{\graph^l\}_{l=0}^\infty\) having the graph posterior as stationary distribution. 
Suppose we have a realisation of length \(M\) of such chain, then the posterior  probability of an edge \(e\) is estimated by \(\frac{1}{M+1-b} \sum_{l=b}^{M} \mathbf{1}_{e}(e^l)\), where the first samples up to \(b\)  are disregarded as a burn-in period. 

The \texttt{mcmc\_heatmaps} module has a list of objects, where each object has an \texttt{id} field for the algorithm object \texttt{id} and a burn-in field (\texttt{burn\_in}) for specifying the burn-in period. 
The estimated probabilities are plotted in heatmaps using \pkg{seaborn} \citep{Waskom2021}.
\\\\
\noindent \emph{MCMC trajectory plots} (\texttt{mcmc\_traj\_plots}) \vspace{0.2cm}\\
The \texttt{mcmc\_traj\_plots} module plots the value of a given functional for the graphs in the MCMC trajectory. 
The currently supported functionals are the number of edges for the graphs (\vals{size}) and the score (\vals{score}).
The \texttt{mcmc\_traj\_plots} module has a list of objects, where each object has an \texttt{id} field for the algorithm object \texttt{id}, a burn-in field (\texttt{burn\_in}) and a field specifying the functional to be considered (\texttt{functional}).
Since the trajectories may be very long, the user may choose to thin out the trajectory by only considering every graph at a given interval length specified by the \texttt{thinning} field.
\\\\
\noindent \emph{MCMC auto-correlation plots} (\texttt{mcmc\_autocorr\_plots})\vspace{0.2cm}\\
The \texttt{mcmc\_autocorr\_plots} module plots the auto-correlation of a functional of the graphs in a MCMC trajectory.
Similar to the \texttt{mcmc\_traj\_plots} module, the  \texttt{mcmc\_autocorr\_plots} module has a list of objects, where each object has an \texttt{id}, \texttt{burn\_in}, \texttt{thinning}, and a \texttt{functional} field. 
The maximum number of lags after thinning, is specified by the \texttt{lags} field.
\\\\

\vskip 0.2in

\bibliography{allbib}

\end{document}